\documentclass{article}
\usepackage{iclr2027/iclr2027_conference,times}
\usepackage{algorithm}
\usepackage{algpseudocode}
\usepackage{amsmath}
\usepackage{amsfonts}
\usepackage{amssymb}
\usepackage{amsthm}
\usepackage{multirow}
\usepackage[utf8]{inputenc}
\usepackage[T1]{fontenc}
\usepackage{url}
\usepackage{booktabs}
\usepackage{microtype}
\usepackage[table]{xcolor}
\usepackage{graphicx}
\usepackage{caption}
\usepackage{placeins}
\usepackage{flafter}
\usepackage{tabularx}
\usepackage{hyperref}
\hypersetup{hidelinks}

\newcolumntype{C}[1]{>{\centering\arraybackslash}p{#1}}
\newcolumntype{Y}{>{\centering\arraybackslash}X}
\theoremstyle{plain}

\begin{document}

\title{When Rerankers Work Together: Better and Faster Reranking without Joint Training}
\author{Yanqiao Chen$^{*\dagger}$, Dongsheng Hou$^{*}$, Yuhan Rui$^{*}$\\[-1pt]
{\normalfont Southern University of Science and Technology}\\[-1pt]
{\normalfont\small \texttt{\{12412115, 12410421\}@mail.sustech.edu.cn}, \texttt{ruiyuhan0110@gmail.com}}\\[-1pt]
{\normalfont\footnotesize $^{*}$Equal contribution.\quad $^{\dagger}$Corresponding author.}\\[-12pt]
}
\iclrfinalcopy
\maketitle

\begin{abstract}
Standalone reranker quality does not determine its value in a multistage RAG pipeline. We systematically study heterogeneous rerankers composed without joint training, spanning lexical, single-vector, late-interaction, and joint query--context scoring. Two lightweight semantic coarse scorers, Cache-Pointwise and Cache-Shapley, learn from passage relevance labels and share cacheable representations and linear-time singleton inference. Under matched candidates and budgets, we compare cascades with stage-removal references to measure evidence retention, repair and breakage, reader accuracy, and online pruning cost. On NoLiMa 32K with Gemma, inserting these scorers before CrossEncoder or Provence improves answer accuracy by 4.02--9.43 percentage points while accelerating online pruning by 1.57--1.88$\times$ relative to the corresponding reranker alone. Cache-Pointwise composed with ColBERTv2 through reciprocal rank fusion improves accuracy by 4.60 points at 1.36$\times$ speedup over direct ColBERTv2. The stage-removal comparisons reveal an asymmetry: a cascade can outperform its final reranker alone while underperforming its coarse selector alone. Two readers, a second multi-hop dataset, and three-stage extensions show that the gains depend on the reader, reranker pairing, and stage configuration. These findings demonstrate a practical route to better and faster reranking without joint training and motivate studying stage order, candidate budgets, and rank fusion as determinants of composition quality.
\end{abstract}

\section{Introduction}

Retrieval-Augmented Generation (RAG) \citep{lewis2021retrievalaugmentedgenerationknowledgeintensivenlp} depends on which retrieved items reach the reader. Large candidate pools improve coverage. They also increase irrelevant text and reranker computation, motivating intermediate selection that preserves required evidence.

RAG commonly combines BM25 or dual-encoder retrieval \citep{robertson2009bm25,karpukhin2020dpr,reimers2019sentencebertsentenceembeddingsusing} with a CrossEncoder \citep{nogueira2020passagererankingbert}. Multi-stage systems already establish the coarse-to-fine topology and its quality--cost trade-off: Feature-based learning-to-rank reduces BERT load \citep{zhang2021muppets}, CMC reranks cached embeddings before a CrossEncoder \citep{song2024cmc}, and FunnelRAG varies granularity, scale, and capacity \citep{zhao2025funnelrag}. Matched-token evaluations show that DOS RAG can match or outperform more complex RAG pipelines \citep{laitenberger2025stronger}.

We study how rerankers with different scoring mechanisms behave when composed without joint training. On NoLiMa \citep{modarressi2025nolima} 32K with Gemma, Cache-Pointwise before CrossEncoder raises answer accuracy from 21.49\% to 28.51\% and accelerates online pruning by 1.58$\times$. Yet Cache-Pointwise alone reaches 31.49\% under the same 30\% token cap (Tables~\ref{tab:nolima_reader_anchors}--\ref{tab:stage_signature_main}). This contrast motivates two questions: when does coarse selection improve an existing reranker, and when does subsequent reranking improve the coarse selector? We test lexical, single-vector, late-interaction, and joint query--context scoring under a shared evaluation protocol.

\begin{figure}[t]
\centering
\includegraphics[width=\linewidth]{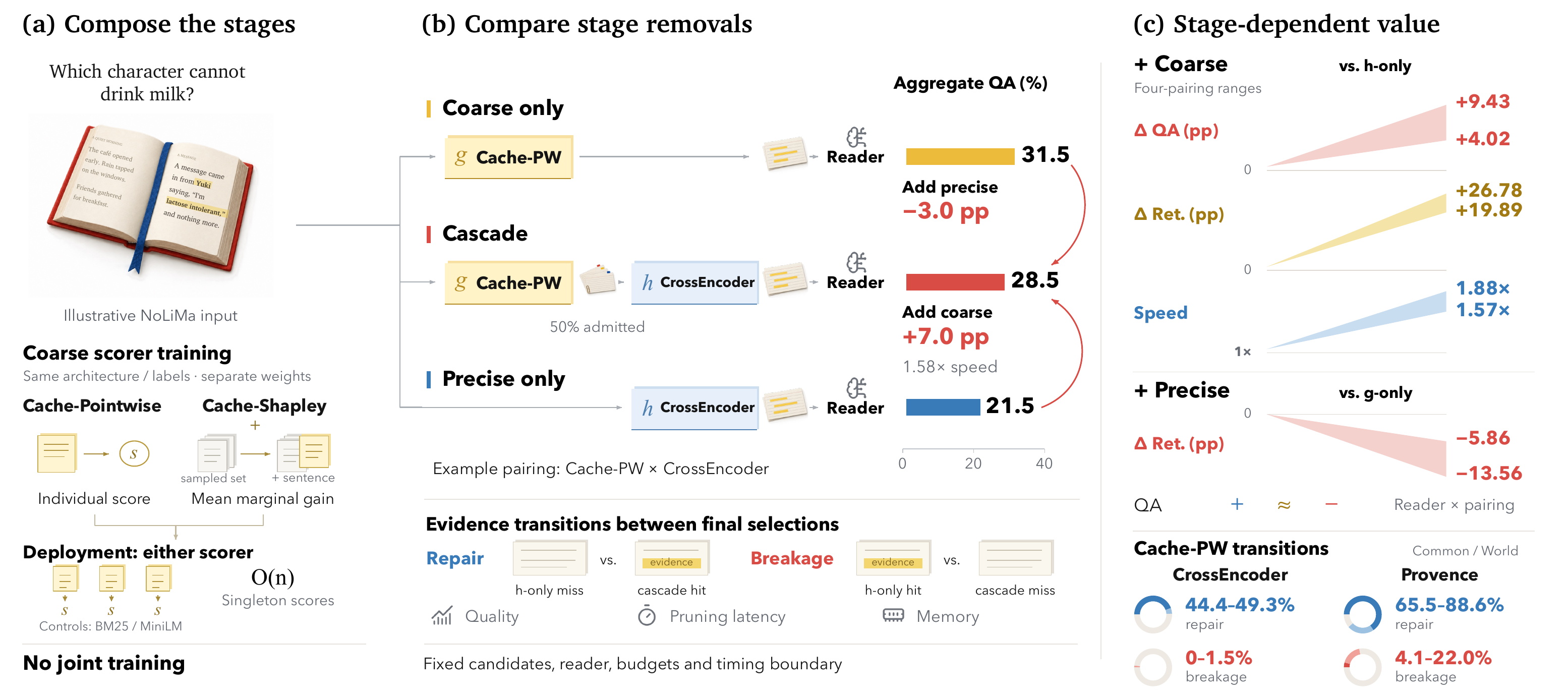}
\caption{Studying heterogeneous rerankers composed without joint training. (A) Lightweight semantic coarse scorers feed a fixed precise reranker under controlled candidates and budgets. (B) Both stage-removal references separate the effects of coarse selection and subsequent reranking, with evidence repair and breakage reported alongside quality and cost. (C) Coarse selection improves precise-only quality and speed together, while the value of subsequent stages depends on the reader and pairing.}
\label{fig:teaser_new}
\end{figure}

To instantiate lightweight semantic coarse selection, our weakly supervised scorer encodes corpus units independently and each query once, then applies a compact set model to reusable pair features. MS MARCO \citep{bajaj2016ms} provides passage-level labels, so each passage remains one training player. We train the coarse scorer separately and compose it with fixed downstream checkpoints, including CrossEncoder and Provence \citep{chirkova2025provence}. No downstream scores, reader feedback, or gradients across stages enter coarse training. This setup tests whether separately developed scoring mechanisms can work effectively together through candidate selection alone.

The matched models differ only in credit training. \textbf{Cache-Pointwise} uses singleton scores in the pairwise loss. \textbf{Cache-Shapley} is a Shapley-trained singleton scorer: it compares sampled-coalition marginal credits during training, then uses the same $O(n)$ singleton readout. Together they test whether composition gains hold across two training objectives under a shared deployment interface. Their matched comparison also isolates how coalition-derived supervision affects coarse-scorer quality and stability.

We cross no filtering, BM25, frozen MiniLM, Cache-Pointwise, and Cache-Shapley with fixed CrossEncoder and Provence stages, and extend the study to ColBERTv2 endpoints and three-stage cascades. Two open-weight readers, knowledge slices, a second multi-hop dataset, Monte-Carlo sweeps, and three seeds establish where these effects persist and where they change.

Our contributions are:

\begin{enumerate}
    \item A systematic study of heterogeneous rerankers composed without joint training, spanning lexical, single-vector, late-interaction, and joint scoring in two- and three-stage pipelines.
    \item Two cacheable semantic coarse rerankers with matched architectures and $O(n)$ singleton deployment, demonstrating that lightweight coarse selection can improve precise-only evidence retention, answer quality, and speed simultaneously.
    \item Controlled stage-removal and evidence-transition analyses that identify improvement, preservation, and degradation across compositions, and establish stage interactions as a design consideration for multistage pruning.
\end{enumerate}

The central finding is that independently trained rerankers can improve quality and speed together, but each stage's value depends on its specific composition.

\section{Related Work}

\paragraph{Multi-stage retrieval and evaluation.} BM25, dual encoders, ColBERTv2 late interaction, and CrossEncoders provide lexical, single-vector, token-level, and joint scoring \citep{robertson2009bm25,karpukhin2020dpr,reimers2019sentencebertsentenceembeddingsusing,santhanam2022colbertv2,nogueira2020passagererankingbert}. Feature-based learning-to-rank filters candidates before BERT \citep{zhang2021muppets}; CMC, CODER, and EBCAR rerank cached embeddings \citep{song2024cmc,zerveas2022coder,yuan2026ebcar}; and FunnelRAG varies granularity and capacity \citep{zhao2025funnelrag}. Evaluations with matched token budgets find strong performance from source-order-preserving RAG relative to summarization-based pipelines \citep{laitenberger2025stronger}. Our protocol compares the same cascade with both stage-removal references at fixed budget and reader.

\paragraph{RAG reranking and context selection.} RECOMP selects or summarizes evidence \citep{xu2023recompimprovingretrievalaugmentedlms}; Selective Context prunes redundant lexical units \citep{li2023compressing}, while LLMLingua and LLMLingua-2 compress prompts \citep{jiang2023llmlingua,pan2024llmlingua2}; and Provence and EXIT prune sentences \citep{chirkova2025provence,hwang2025exit}. LooComp learns sentence-removal scores from evidence annotations \citep{do2026loocompleverageleaveoneoutstrategy}. Other methods optimize masks or budgets \citep{huang2025gumbel,taguchi2025adaptivek,meng2025dps}. Unlike direct selection, our coarse output feeds a fixed precise stage, exposing both admitted work and downstream compatibility.

\paragraph{Context value and generator feedback.} Leave-one-out attribution and influence-based selection quantify context utility relative to a target generator \citep{liu2025attribot,deng2025influence}; attention-based attribution learns to approximate ablation effects \citep{cohenwang2025attributionattention}. DynamicRAG and BAR-RAG optimize evidence selection using generator feedback \citep{sun2025dynamicrag,sun2026barrag}. Our training uses passage relevance for reader-independent reuse, and fixed LLMs supply downstream evaluation.

\paragraph{RAG evaluation frameworks.} RAGAS and RAGVUE decompose RAG outputs into retrieval, answer, and grounding dimensions \citep{es2024ragas,murugaraj2026ragvue}. RAG Playground compares retrieval strategies and prompting choices \citep{papadimitriou2024ragplayground}; \citet{simon2024ragevaluation} propose a systematic evaluation methodology with a configuration-dependency case study. SCARF evaluates deployed RAG configurations through a black-box interface \citep{rengo2025scarf}. These frameworks assess pipeline outputs or compare complete configurations. Our protocol attributes changes inside a reranking cascade through matched stage-removal references, conditional evidence transitions, and explicit online cost coordinates.

\section{Composing Heterogeneous Rerankers}\label{sec:method}

We compose stages with different scoring mechanisms while keeping their parameters fixed during cascade execution. Learned coarse scorers are trained separately from the downstream rerankers and reader; the cascade has no joint training objective. The stages interact through the candidates and rankings passed between them. This design lets us study how composition changes quality and cost using existing reranking components.

\subsection{Evaluation}\label{sec:cascade_role}

For a query $q$ and retrieved candidate units $C=(c_1,\ldots,c_n)$, a cascade is specified by $\mathcal F=(g,\alpha,h,B,\mathcal R)$: coarse ranker $g$, admitted fraction $\alpha$, precise reranker $h$, final context budget $B$, and reader $\mathcal R$. Candidate granularity follows the dataset: MS MARCO passages during training, HotpotQA \citep{yang2018hotpotqa} and 2WikiMultiHopQA \citep{ho2020constructing} sentences, MuSiQue \citep{trivedi2022musique} paragraphs, and NoLiMa sentences. NoLiMa measures $B$ in tokens; 2WikiMultiHopQA measures it in sentences. The execution path is
\[
C_1=\operatorname{Top}_{k}(C;g), \qquad
C_2=\operatorname{BudgetTop}_{B}(C_1;h), \qquad
\hat a=\mathcal R(q,C_2).
\]
Here $k=\min(n,\max(1,\operatorname{round}(\alpha n)))$ for nonempty inputs. We record both the evidence surviving in $C_1$ and the candidate load processed by $h$. Direct coarse deployment applies $\operatorname{BudgetTop}_{B}(C;g)$ to the full candidate pool. Selected sentence-level units return to source order before generation.

For each cascade, we record the outcomes $\mathbf y(\mathcal F)=(\mathrm{Rank},\mathrm{Ret}_1,\mathrm{Ret}_2,\mathrm{QA},T_{\mathrm{online}},M_{\mathrm{peak}})$. These measure candidate ordering, coarse and final evidence survival, reader behavior, synchronized latency, and memory. Within each cascade panel, we hold $(\alpha,h,B,\mathcal R)$ fixed and vary $g$. BM25 supplies a lexical control, frozen MiniLM a cacheable semantic control, and the learned scorers test weak supervision under the same interface.

\paragraph{Two stage-removal comparisons.} Let $Y_m(g{\rightarrow}h;\alpha,B,\mathcal R)$ denote the paired mean of outcome $m$ for the cascade, $Y_m(h;B,\mathcal R)$ the precise-only reference, and $Y_m(g;B,\mathcal R)$ the coarse-only reference. We report
\[
\Delta_g^m = Y_m(g{\rightarrow}h)-Y_m(h),\qquad
\Delta_h^m = Y_m(g{\rightarrow}h)-Y_m(g).
\]
$\Delta_g^m$ measures coarse-stage insertion relative to precise-only deployment; $\Delta_h^m$ measures subsequent reranking relative to coarse-only deployment. Ret denotes final evidence survival under budget $B$; positive retention or QA contrasts indicate improvement. For cost, we report speedup $\rho_T=\overline{T}(h)/\overline{T}(g{\rightarrow}h)$, the ratio of mean online latencies, and memory saving $\Delta M=M(h)-M(g{\rightarrow}h)$, where $M$ is the maximum allocated GPU-memory peak across requests. We collect these comparisons in
\[
\boldsymbol\sigma(g,h,\mathcal R)=
(\Delta_g^{\mathrm{Ret}},\Delta_h^{\mathrm{Ret}},
\Delta_g^{\mathrm{QA}},\Delta_h^{\mathrm{QA}},\rho_T,\Delta M)
\]
and report each quantity separately. Rank and $\mathrm{Ret}_1$ remain stage diagnostics; CERR and CEBR below describe paired evidence transitions.

\paragraph{Cascade evidence repair and breakage.} For binary final evidence survival $E$ and a reference system $b$, we further report the cascade evidence repair rate (CERR) and cascade evidence breakage rate (CEBR):
\[
\mathrm{CERR}_b=\Pr(E_{g\to h}{=}1\mid E_b{=}0),\quad
\mathrm{CEBR}_b=\Pr(E_{g\to h}{=}0\mid E_b{=}1).
\]
CERR measures the fraction of reference misses repaired by the cascade---an operational evidence-complementarity rate; CEBR measures damaged reference hits. They satisfy $\Delta_b^{\mathrm{Ret}}=(1-\mathrm{Ret}_b)\mathrm{CERR}_b-\mathrm{Ret}_b\mathrm{CEBR}_b$. We report the pair to expose both evidence recovery and disruption. Section~\ref{sec:knowledge_slices} uses the precise-only reference $b=h$.

\begin{figure}[t]
\centering
\includegraphics[width=\linewidth]{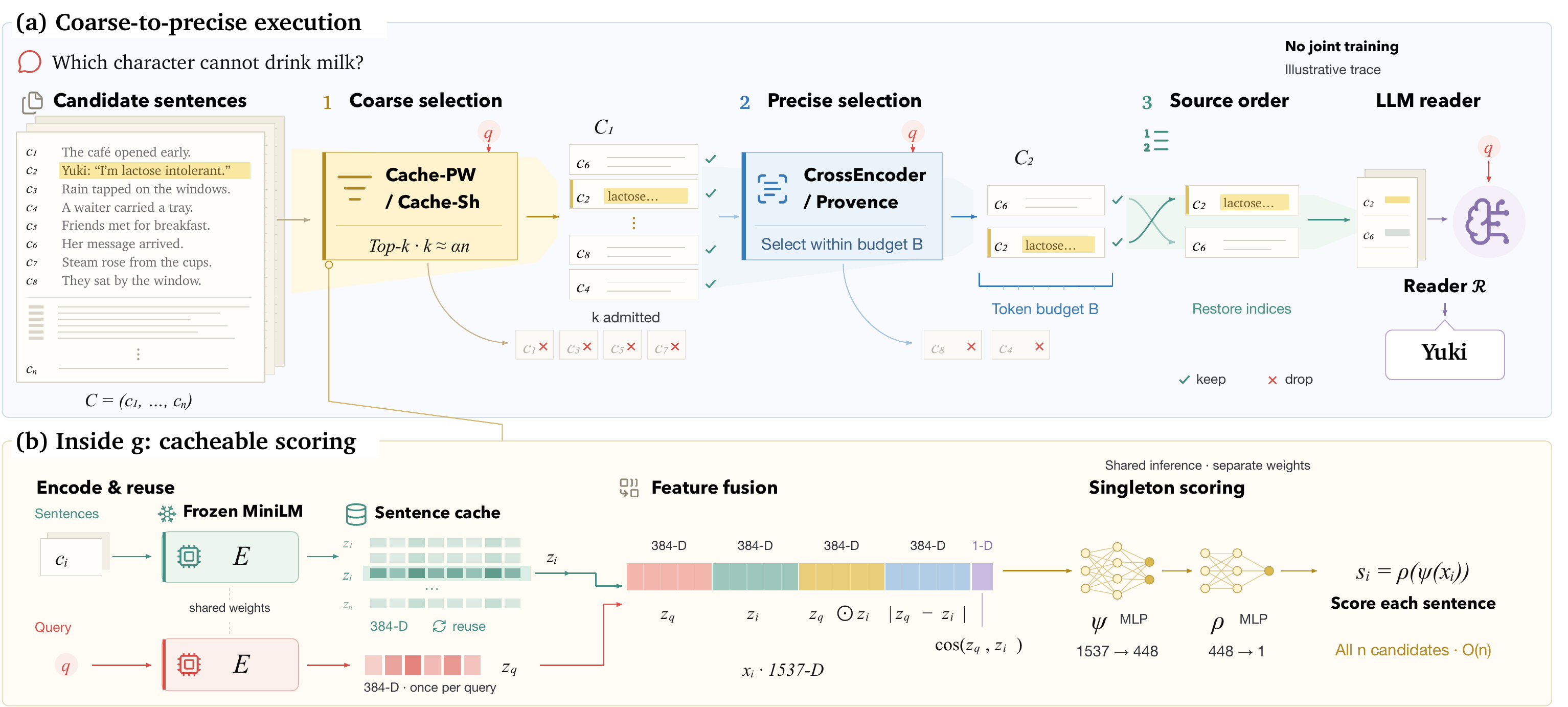}
\caption{Execution path of the evaluated cascade. The cacheable semantic coarse ranker scores the complete retrieved context, the precise stage receives a budgeted candidate set, and the final context is restored to source order for the reader.}
\label{fig:cascade_position}
\end{figure}

\subsection{Cacheable Semantic Scorer}

We use a frozen SentenceTransformer as the encoder $E$. Corpus units are encoded independently and stored as normalized vectors $z_i=E(c_i)$. The online query is encoded once as $z_q=E(q)$. For each pair we construct $x_i=[z_q;z_i;z_q\odot z_i;|z_q-z_i|;\cos(z_q,z_i)]$, 1,537-dimensional for the MiniLM encoder \citep{wang2020minilmdeepselfattentiondistillation} used in our experiments. A learned transformation $\psi:\mathbb{R}^{1537}\rightarrow\mathbb{R}^{448}$ maps $x_i$ to a latent vector. For a nonempty candidate set $S$, a Deep Sets scorer \citep{zaheer2018deepsets,wagstaff2019limitations} computes
\[
v_\theta(S)=\rho\!\left(\sum_{i\in S}\psi(x_i)\right),
\qquad v_\theta(\emptyset)=0,
\]
where $\rho:\mathbb{R}^{448}\rightarrow\mathbb{R}$ is a scalar-output multilayer perceptron. The two networks contain 3,394,497 trainable parameters in total. The encoder remains frozen. The corpus cache stores $z_i$ and permits reuse across queries; query-dependent features $x_i$ are assembled online. This design learns query-dependent relevance scores while reusing the same frozen corpus representations across queries.

\subsection{Pointwise and Shapley Credit Training}

Each training context provides relevant-passage indices $\mathcal{P}$ and nonselected-passage indices $\mathcal{D}$. Both objectives optimize a pairwise margin $\epsilon$ between positive and negative candidate credits:
\[
\mathcal{L}(\theta)=\frac{1}{|\mathcal{P}||\mathcal{D}|}
\sum_{i\in\mathcal{P}}\sum_{j\in\mathcal{D}}
\max\{0,\epsilon-(r_i-r_j)\}.
\]

\paragraph{Pointwise credit.} Cache-Pointwise sets $r_i=v_\theta(\{c_i\})$. This objective learns an independent score for each query--candidate feature under the shared scorer.

\paragraph{Shapley credit.} Following SCP \citep{chen2026shapleycontextpruningcooperative}, Cache-Shapley views the candidate set as a cooperative game with scalar characteristic function $v_\theta$ \citep{shapley1953value}. The exact Shapley credit of candidate unit $i$ is
\[
\phi_i(v_\theta)=
\sum_{S\subseteq C\setminus\{i\}}
\frac{|S|!(n-|S|-1)!}{n!}
\left[v_\theta(S\cup\{i\})-v_\theta(S)\right].
\]
We estimate $\phi_i$ from $M$ uniformly sampled permutations \citep{castro2009polynomial}. Each candidate unit receives the average marginal score observed when it enters a permutation prefix. Cache-Shapley uses these estimates as $r_i$ during training, with $M=10$; its name refers to this supervision rule. The matched objectives therefore isolate training-credit choice while holding the representation, scorer architecture, and passage labels fixed.

\subsection{Singleton Deployment and Online Cost}

Both cacheable models deploy through singleton scores $r_i=v_\theta(\{c_i\})$. Each query requires one query encoding, $n$ feature constructions, and $n$ compact scorer evaluations; corpus-unit encoding occurs once when the cache is populated. The precise reranker then processes the admitted candidates together with the query. Cache-Shapley uses Shapley credits exclusively as its training supervision rule; deployment uses the learned singleton value $v_\theta(\{c_i\})$ as a relevance score.

\subsection{Fixed-Budget Selection}

The experiments sort candidates by singleton score, apply the dataset's count or token budget, and restore source order. Count selection retains $\operatorname{round}(rn)$ units, clipped to $[1,n]$ for nonempty inputs, with ties rounded to even. NoLiMa greedily scans descending scores, skips units that do not fit the remaining token budget, and continues to lower-ranked units. It then counts the joined context exactly and removes the lowest-scored selected units if needed to meet the cap. Appendix~\ref{sec:algorithms_new} gives complete pseudocode for matched training and semantic cascade inference.

\section{Experiments}\label{sec:experiments}

\subsection{Experimental Setup}

Cache-Pointwise and Cache-Shapley share a frozen MiniLM-L6-v2 encoder, scorer architecture, MS MARCO passage supervision, optimization, and singleton deployment; only the training credit differs. We evaluate passage ranking, zero-shot sentence and paragraph ranking, NoLiMa 32K selection with Gemma 3 \citep{gemmateam2025gemma3} and Llama 3.1 \citep{grattafiori2024llama3} readers, and cascades ending in CrossEncoder, ColBERTv2, or Provence. Every cascade uses matched examples, fixed stage loads, and a common final budget. We report ranking, exact evidence retention, reader quality, synchronized online latency, and memory.

The experiments test coarse-scorer transfer (Section~\ref{sec:cache_ranking}), coarse-only reader outcomes (Section~\ref{sec:nolima_direct}), and heterogeneous cascades (Section~\ref{sec:cascade_results}). Sections~\ref{sec:knowledge_slices}--\ref{sec:2wiki_results} examine variation across knowledge types and datasets. Appendix~\ref{app:exp_results} gives the full evaluation and execution setup. Appendix~\ref{sec:mc_noise_supp} reports training-credit, readout, and multi-seed diagnostics.

\subsection{Pre-Composition Transfer Across Candidate Granularities}\label{sec:cache_ranking}

\paragraph{Passage supervision transfers across candidate granularities.} This experiment evaluates the coarse scorer before any precise reranker or reader is introduced. Frozen cosine and the learned heads share the same candidate embeddings, so their difference isolates the learned query--candidate feature fusion and its weak supervision. Both learned scorers improve frozen MiniLM cosine on MS MARCO, HotpotQA, and 2WikiMultiHopQA across NPRS and standard ranking metrics. Frozen cosine leads on MuSiQue. A matched BGE control reproduces the same three-positive/one-negative transfer pattern.

Appendix Tables~\ref{tab:cache_standard_metrics}--\ref{tab:bge_control} contain the complete NPRS, MAP, nDCG, evidence-recall, and paired results. Passage labels thus support transfer to sentence candidates, while MuSiQue shows that this advantage varies across target tasks.

\subsection{Coarse-Only References Separate Evidence from Reader Response}\label{sec:nolima_direct}

\paragraph{Retained evidence and surrounding text jointly shape reader input.} NoLiMa removes literal question--evidence overlap and evaluates semantic selection under identical 32K contexts and token budgets. A selector determines whether the annotated evidence survives and which surrounding sentences reach the reader; every method restores source order before generation. Table~\ref{tab:nolima_reader_anchors} presents evidence-preserving reader anchors alongside key direct and cascade results. Appendix Table~\ref{tab:nolima_direct} supplies the complete direct reference panel.

\begin{table}[t]
\centering
\small
\setlength{\tabcolsep}{7pt}
\caption{Reader anchors and core NoLiMa 32K results over 870 matched contexts. Ret. is exact evidence retention; Tok. is the fraction of original tokens delivered to the reader. QA-G/L denote Gemma/Llama answer accuracy; -- denotes unreported anchor results. Selection rows share a 30\% token cap.}
\label{tab:nolima_reader_anchors}
\begin{tabular}{lcccc}
\toprule
\textbf{Reader input / selector} & \textbf{Ret.} & \textbf{Tok.} & \textbf{QA-G} & \textbf{QA-L} \\
\midrule
Full 32K context & 1.0000 & 1.0000 & 0.2218 & -- \\
Oracle evidence only & 1.0000 & 0.0008 & 0.9506 & -- \\
\midrule
CrossEncoder & 0.5069 & 0.2752 & 0.2149 & 0.1966 \\
Provence & 0.5034 & 0.2840 & 0.2230 & 0.1644 \\
Frozen MiniLM & 0.7218 & 0.2850 & 0.2770 & 0.2149 \\
Cache-Shapley & 0.7644 & 0.2844 & 0.3034 & 0.2069 \\
Cache-Pointwise & \textbf{0.8609} & 0.2849 & 0.3149 & \textbf{0.2655} \\
Cache-PW$\rightarrow$CrossEncoder & 0.7253 & 0.2814 & 0.2851 & 0.2448 \\
Cache-PW$\rightarrow$Provence & 0.7713 & 0.2845 & \textbf{0.3172} & 0.2184 \\
\bottomrule
\end{tabular}
\end{table}

Both anchors retain all gold evidence, yet removing the surrounding book raises QA from 22.18\% to 95.06\%. Evidence retention therefore leaves substantial variation in reader accuracy unexplained. An evidence-fixed distractor control in Appendix~\ref{sec:evidence_fixed_supp} further examines this effect.

Among direct selectors, Cache-Pointwise achieves the strongest aggregate retention and QA for both readers; Cache-Shapley improves the shared frozen representation with Gemma. The Cache-PW--Provence cascade reaches 31.72\% QA with Gemma but 21.84\% with Llama, below direct Cache-PW's 26.55\% for that reader. Appendix Tables~\ref{tab:nolima_direct} and~\ref{tab:nolima_books_full_new}--\ref{tab:nolima_books_llama31_8b} give complete method, book, and uncertainty results.

\subsection{Heterogeneous Cascades Improve Quality and Speed}\label{sec:cascade_results}

Table~\ref{tab:stage_signature_main} reports both stage-removal contrasts defined in Section~\ref{sec:cascade_role}. Appendix Table~\ref{tab:nolima_cascade} gives absolute outcomes, and Appendix Tables~\ref{tab:stage_signature_ci}--\ref{tab:coarse_load_sensitivity} provide uncertainty, timing, and budget decompositions.

\begin{table}[t]
\centering
\small
\setlength{\tabcolsep}{8pt}
\caption{Stage-removal comparisons on NoLiMa with Gemma, in percentage points except $\rho_T$ and GiB. $\Delta_g$ compares the cascade with the precise-only row; $\Delta_h$ compares it with the coarse-only row. Positive retention/QA contrasts, speedup above one, and positive $\Delta M$ are favorable. Values use unrounded paired outcomes; $\rho_T$ is the ratio of mean latencies.}
\label{tab:stage_signature_main}
\begin{tabular}{llcccccc}
\toprule
$g$ & $h$ & $\Delta_g^{\rm Ret}$ & $\Delta_h^{\rm Ret}$ & $\Delta_g^{\rm QA}$ & $\Delta_h^{\rm QA}$ & $\rho_T$ & $\Delta M$ \\
\midrule
Cache-PW & CrossEncoder & +21.84 & $-13.56$ & +7.01 & $-2.99$ & 1.58 & $-0.56$ \\
Cache-Sh. & CrossEncoder & +19.89 & $-5.86$ & +4.02 & $-4.83$ & 1.57 & $-0.56$ \\
Cache-PW & Provence & +26.78 & $-8.97$ & +9.43 & +0.23 & 1.86 & +12.28 \\
Cache-Sh. & Provence & +19.89 & $-6.21$ & +6.67 & $-1.38$ & 1.88 & +12.43 \\
\bottomrule
\end{tabular}
\end{table}

Every learned coarse stage improves evidence retention, QA, and speed relative to its precise-only endpoint. This positive $\Delta_g$ means that semantic coarse selection gives the precise reranker a smaller and more useful candidate distribution. The simple pointwise scorer raises CrossEncoder retention by 21.84 points and QA by 7.01 points at 1.58$\times$ speed, with both models trained separately. Thus, coarse pruning can improve what a fixed downstream reranker selects while reducing how much it processes. BM25 reduces computation and lowers final quality, demonstrating that effective coarse selection requires candidate quality.

The reverse comparison is significantly negative in retention for every precise stage. Both CrossEncoder pairings also show significantly negative QA contrasts, while the Provence QA contrasts do not differ significantly from zero. A negative $\Delta_h^{\rm Ret}$ means that the precise reranker removes evidence preserved by the coarse ranker under the same final budget. A negative $\Delta_h^{\rm QA}$ additionally captures changes in retained distractors and reader response. The same cascade can therefore improve a precise-only system while its later stage reduces coarse-only quality. These comparisons support both improving an existing precise system and deciding which stages to retain when the architecture is flexible (Appendix~\ref{sec:multistage_implications}). Llama reproduces the positive coarse-stage result and changes the precise-stage ordering. Because the selected contexts stay fixed across readers, this change isolates reader identity as a component of the composition.

\paragraph{Extending the cascade to ColBERTv2.} CrossEncoder jointly encodes each query--sentence pair; Provence scores sentences within the supplied query-conditioned context. ColBERTv2 scores query--sentence pairs through token-level matching, using sentence representations that can be cached and reused. We instantiate it as a final endpoint and as a middle stage before the two joint rerankers. Hard fusion uses the ColBERTv2 order inside the admitted set; Reciprocal rank fusion (RRF) \citep{cormack2009rrf} combines upstream and ColBERTv2 ranks. This comparison tests whether the framework applies across scoring interfaces and whether upstream rank information should be discarded or fused. Table~\ref{tab:colbert_endpoint_main} reports the endpoint comparison. Appendix Tables~\ref{tab:colbert_selection_panel}--\ref{tab:colbert_depth_ci} report the complete two- and three-stage panel with matched removal references and paired intervals.

\begin{table}[t]
\centering
\small
\setlength{\tabcolsep}{7pt}
\caption{ColBERTv2 as the final reranker on NoLiMa 32K. Load is the fraction of the original sentence pool scored by ColBERTv2. Retention and Gemma QA use the same 870 contexts and common 30\% final token cap; speedup is the ratio of mean latencies relative to direct ColBERTv2. RRF uses $k{=}60$.}
\label{tab:colbert_endpoint_main}
\begin{tabular}{lcccccc}
\toprule
\textbf{Upstream stage} & \textbf{Fusion} & \textbf{Load} & \textbf{p50 ms} & \textbf{Speedup} & \textbf{Retention} & \textbf{QA} \\
\midrule
None (direct ColBERTv2) & -- & 100\% & 97.7 & 1.00$\times$ & 0.7897 & 0.2954 \\
Random & Hard & 50\% & 59.9 & 1.71$\times$ & 0.4701 & 0.1943 \\
BM25 & Hard & 50\% & 72.6 & 1.40$\times$ & 0.5805 & 0.2563 \\
Frozen MiniLM & Hard & 50\% & 68.0 & 1.43$\times$ & 0.8149 & 0.3172 \\
Cache-Shapley & Hard & 50\% & 72.6 & 1.35$\times$ & 0.8391 & 0.3230 \\
Cache-Pointwise & Hard & 50\% & 83.8 & 1.18$\times$ & 0.9149 & \textbf{0.3414} \\
Cache-Pointwise & Hard & 70\% & 87.6 & 1.13$\times$ & 0.8667 & 0.3310 \\
Cache-Shapley & RRF & 50\% & 73.1 & 1.35$\times$ & 0.8529 & 0.3368 \\
Cache-Pointwise & RRF & 50\% & 72.7 & 1.36$\times$ & \textbf{0.9230} & \textbf{0.3414} \\
\bottomrule
\end{tabular}
\end{table}

\paragraph{Learned coarse selection improves ColBERTv2.} Random and BM25 coarse selection reduce ColBERTv2 retention and QA. Frozen MiniLM improves both retention and QA, and the learned stages provide the strongest candidate distributions. Cache-Pointwise with RRF reaches 0.9230 retention and 0.3414 QA and runs 1.36$\times$ faster than direct ColBERTv2. Cache-Shapley with RRF reaches similar QA with lower retention. Hard and RRF fusion produce the same Cache-Pointwise QA through different final rankings. The gains with both Hard and RRF show that learned coarse selection can improve ColBERTv2 under either ranking strategy.

\paragraph{Depth extension at a fixed load reveals composition-specific value.} At the tested 70\% ColBERTv2 and 50\% endpoint loads, ColBERTv2 before CrossEncoder or Provence improves both direct precise endpoints. A preceding Cache-Pointwise stage raises the ColBERTv2--Provence QA from 0.2747 to 0.3218 and leaves the ColBERTv2--CrossEncoder QA essentially unchanged. The third stage therefore adds reader value in one chain and primarily changes workload and evidence flow in the other. The matched removal panel identifies these roles by deleting one stage at a time. These roles do not transfer across rerankers or operating points.

\subsection{Compatibility Varies Across Knowledge Regimes}\label{sec:knowledge_slices}

\paragraph{Commonsense and world knowledge reward different candidate distributions.} An aggregate score can hide a gain on one reasoning regime and a loss on another. Table~\ref{tab:knowledge_slices_main} therefore separates NoLiMa outcomes by knowledge type under identical inputs and budgets. Accuracy measures reader response and retention measures evidence survival. Repair/breakage (R/B) compares each cascade with its precise-only reference: repair is the fraction of reference misses recovered by the cascade, and breakage is the fraction of reference hits lost by the cascade. These conditional transitions explain how the net retention difference is formed. Appendix Tables~\ref{tab:knowledge_slices_ci} and~\ref{tab:evidence_repair_ci} provide complete intervals and denominators.

\begin{table}[t]
\centering
\scriptsize
\setlength{\tabcolsep}{1.5pt}
\caption{NoLiMa 32K knowledge-type results with Gemma 3 12B, in percent. Common and World contain 240 and 630 outcomes. Acc. is answer accuracy; Ret. is exact evidence retention; R/B is CERR/CEBR against the corresponding precise-only system.}
\label{tab:knowledge_slices_main}
\begin{tabular}{C{2.05cm}C{1.25cm}C{1.25cm}C{1.25cm}C{1.25cm}C{1.75cm}C{1.75cm}}
\toprule
\textbf{Method} & \textbf{Common Acc.} & \textbf{Common Ret.} & \textbf{World Acc.} & \textbf{World Ret.} & \textbf{Common R/B} & \textbf{World R/B} \\
\midrule
\multicolumn{7}{c}{\textit{Baselines and precise rerankers}} \\[2.5pt]
Random & 20.00 & 27.08 & 11.27 & 29.05 & -- & -- \\
BM25 & 37.50 & 49.17 & 7.14 & 24.92 & -- & -- \\
LLMLingua2 & 27.92 & -- & 11.59 & -- & -- & -- \\
CrossEncoder & 45.42 & 70.42 & 12.38 & 43.17 & -- & -- \\
Provence & 42.92 & 70.83 & 14.44 & 42.54 & -- & -- \\
\midrule
\multicolumn{7}{c}{\textit{Cacheable direct methods and cascades}} \\[2.5pt]
Cache-Shapley & 51.25 & 95.00 & 22.38 & 69.37 & -- & -- \\
Cache-Pointwise & 49.17 & \textbf{95.42} & \textbf{24.76} & \textbf{82.54} & -- & -- \\
Cache-Sh.$\rightarrow$CE & 46.67 & 85.42 & 17.46 & 64.92 & 50.70/0.00 & 43.30/6.62 \\
Cache-PW$\rightarrow$CE & 47.92 & 85.00 & 21.11 & 67.78 & 49.30/0.00 & 44.41/1.47 \\
Cache-Sh.$\rightarrow$Prov. & 54.17 & 88.33 & 19.37 & 63.33 & 77.14/7.06 & 55.25/25.75 \\
Cache-PW$\rightarrow$Prov. & \textbf{54.58} & 93.75 & 23.02 & 70.79 & \textbf{88.57/4.12} & \textbf{65.47/22.01} \\
\bottomrule
\end{tabular}
\end{table}

\paragraph{Repair and breakage localize evidence complementarity.} The learned cascades repair precise-reranker misses in both knowledge regimes. Provence after Cache-Pointwise repairs 237 of 362 prior misses on world knowledge and 62 of 70 on commonsense: world knowledge has more repaired misses in absolute terms but a lower repair rate and a higher breakage rate. Commonsense QA improves over direct Cache-Pointwise; world-knowledge QA decreases slightly. Both cascade slices exceed direct Provence.

The transition rates show how recovered and newly lost evidence contribute to aggregate retention gains, while QA identifies where the reader benefits from the resulting context.

\subsection{A Second Dataset Separates Improvement from Preservation}\label{sec:2wiki_results}

Table~\ref{tab:2wiki_cascade} evaluates 600 2WikiMultiHopQA questions, admitting 70\% and finally retaining 50\% of the original sentences. Cache-Pointwise raises Provence's F1 by 0.0411 (95\% CI [0.0037, 0.0780]); Cache-Shapley attains the highest Provence evidence recall. Semantic coarse stages do not significantly change CrossEncoder F1, whereas BM25 lowers both F1 and evidence recall. Appendix~\ref{sec:2wiki_paired_supp} gives additional paired statistics.

\begin{table}[t]
\centering
\small
\setlength{\tabcolsep}{7pt}
\caption{2WikiMultiHopQA cascade results with Gemma 3 12B. F1 is best-alias token multiset F1 after answer normalization; EvR is supporting-evidence recall. Best values in each precise-reranker panel are bold.}
\label{tab:2wiki_cascade}
\begin{tabular}{lcccc}
\toprule
& \multicolumn{2}{c}{\textbf{CrossEncoder}} & \multicolumn{2}{c}{\textbf{Provence}} \\
\cmidrule(lr){2-3}\cmidrule(lr){4-5}
\textbf{Coarse stage} & \textbf{F1} & \textbf{EvR} & \textbf{F1} & \textbf{EvR} \\
\midrule
None & 0.5584 & 0.8817 & 0.5269 & 0.7800 \\
BM25 & 0.5053 & 0.7692 & 0.5066 & 0.7308 \\
MiniLM & \textbf{0.5658} & 0.8817 & 0.5545 & 0.8321 \\
Cache-Shapley & 0.5458 & 0.8850 & 0.5548 & \textbf{0.8642} \\
Cache-Pointwise & 0.5521 & \textbf{0.8879} & \textbf{0.5680} & 0.8571 \\
\bottomrule
\end{tabular}
\end{table}

\paragraph{Training-seed sensitivity.} Appendix Table~\ref{tab:multiseed_retention} reports NoLiMa retention across three training seeds. Pointwise has the higher mean and maximum; Shapley has the smaller observed range. These diagnostics cover retention only; the reader results above use seed 42.

\section{Limitations and Future Work}

\paragraph{Limitation: scope of the current evidence.} The experiments cover a limited set of scenarios and configurations. Establishing broader generality requires more languages, context lengths, and candidate granularities.

\paragraph{Adaptive pipeline design.} Future pipelines should jointly choose stage order, candidate budgets, and how upstream rankings are passed to later stages. Our three-stage results motivate query-dependent routing: a controller could stop after coarse selection, invoke a further reranker, or preserve upstream scores through rank fusion. Budget allocation should account for both evidence loss and downstream workload, with the reader included in the quality objective. Independently trained compositions also provide a reference for evaluating joint training. Appendix~\ref{sec:multistage_implications} illustrates deployment choices, and Appendix~\ref{sec:research_outlook} develops these directions.

\paragraph{Pipelined serving at corpus scale.} The experiment results show that pipelined reranking is a promising serving design. A serving pipeline must also coordinate encoding, coarse selection, precise reranking, and generation. Stage-local queues could overlap these operations across requests, using separate batch sizes and service rates. Candidate budgets would then affect queueing, accelerator utilization, and tail latency as well as evidence survival. Future studies should jointly measure end-to-end latency, throughput, buffer and cache memory, and QA, including index maintenance and data transfer.

\section{Conclusion}

We systematically studied heterogeneous RAG rerankers composed without joint training across stages. Lightweight semantic coarse selection improves the evidence retention and answer quality of fixed precise rerankers while accelerating online pruning. The gains demonstrate that separately developed scoring mechanisms can form effective cascades through candidate selection and rank fusion. The stage-removal comparisons also show why these gains depend on composition: a cascade can improve its precise-only endpoint while a later stage reduces coarse-only quality. Reader identity, knowledge type, and reranker pairing change this balance. These findings motivate multistage pruning as a research direction, with stage order, candidate budgets, rank fusion, and interactions among scoring mechanisms as central design choices. The evaluation protocol provides a common basis for developing and testing such pipelines.

\clearpage
\subsection*{AI Use Statement}

Generative AI tools assisted with language editing, presentation refinement, preliminary discovery and organization of potentially relevant literature, and limited drafting and debugging of experiment-support code. The authors manually inspected the cited sources, defined the methodology and evaluation protocol, checked implementations and data provenance, executed the experiments, and verified the reported numbers and claims. No AI-assisted material or technical output was accepted without author review. The authors take responsibility for the complete content of this work.

\subsection*{Ethics Statement}

This study evaluates retrieval, reranking, and question-answering systems on released research benchmarks and introduces no new human-subject study or collection of private user data. RAG cascades can omit relevant evidence, amplify biases in retrieved corpora, and produce unsupported answers when deployed beyond the evaluated settings. We therefore report evidence loss together with answer quality, restrict our conclusions to the declared datasets, readers, budgets, and rerankers, and recommend application-specific validation and data governance before deployment.

\subsection*{Reproducibility Statement}

Appendix~\ref{app:exp_results} specifies datasets, splits, candidate granularity, model versions, training settings, stage budgets, evaluation metrics, statistical units, and timing boundaries. Appendix~\ref{sec:algorithms_new} provides the training and inference algorithms. The accompanying artifact contains entry points for training, ranking, NoLiMa, QA, cascade, timing, bootstrap, and result summarization; its records retain per-example identifiers, selected indices, evidence survival, reader outputs, synchronization flags, and random seeds. These materials map each reported result to its configuration and evaluation procedure.

\bibliography{references}

@article{bajaj2016ms,
  author  = {Payal Bajaj and Daniel Campos and Nick Craswell and Li Deng and Jianfeng Gao and Xiaodong Liu and Rangan Majumder and Andrew McNamara and Bhaskar Mitra and Tri Nguyen and Mir Rosenberg and Xia Song and Alina Stoica and Saurabh Tiwary and Tong Wang},
  title   = {{MS MARCO}: A human generated machine reading comprehension dataset},
  journal = {arXiv preprint arXiv:1611.09268},
  year    = {2016},
  url     = {https://arxiv.org/abs/1611.09268}
}

@article{castro2009polynomial,
  author  = {Javier Castro and Daniel G{\'o}mez and Juan Tejada},
  title   = {Polynomial calculation of the {Shapley} value based on sampling},
  journal = {Computers \& Operations Research},
  volume  = {36},
  number  = {5},
  pages   = {1726--1730},
  year    = {2009},
  doi     = {10.1016/j.cor.2008.04.004}
}

@inproceedings{chirkova2025provence,
  author    = {Nadezhda Chirkova and Thibault Formal and Vassilina Nikoulina and St{\'e}phane Clinchant},
  title     = {Provence: Efficient and robust context pruning for retrieval-augmented generation},
  booktitle = {International Conference on Learning Representations (ICLR)},
  year      = {2025},
  note      = {arXiv:2501.16214},
  url       = {https://openreview.net/forum?id=TDy5Ih78b4}
}

@inproceedings{ho2020constructing,
  author    = {Xanh Ho and Anh-Khoa Duong Nguyen and Saku Sugawara and Akiko Aizawa},
  title     = {Constructing a multi-hop {QA} dataset for comprehensive evaluation of reasoning steps},
  booktitle = {Proceedings of the 28th International Conference on Computational Linguistics},
  year      = {2020},
  pages     = {6609--6625},
  publisher = {International Committee on Computational Linguistics},
  doi       = {10.18653/v1/2020.coling-main.580},
  url       = {https://aclanthology.org/2020.coling-main.580/}
}

@inproceedings{jiang2023llmlingua,
  author    = {Huiqiang Jiang and Qianhui Wu and Chin-Yew Lin and Yuqing Yang and Lili Qiu},
  title     = {{LLMLingua}: Compressing prompts for accelerated inference of large language models},
  booktitle = {Proceedings of the 2023 Conference on Empirical Methods in Natural Language Processing (EMNLP)},
  year      = {2023},
  pages     = {13358--13376},
  publisher = {Association for Computational Linguistics},
  doi       = {10.18653/v1/2023.emnlp-main.825},
  url       = {https://aclanthology.org/2023.emnlp-main.825/}
}

@inproceedings{karpukhin2020dpr,
  author    = {Vladimir Karpukhin and Barlas Oguz and Sewon Min and Patrick Lewis and Ledell Wu and Sergey Edunov and Danqi Chen and Wen-tau Yih},
  title     = {Dense passage retrieval for open-domain question answering},
  booktitle = {Proceedings of the 2020 Conference on Empirical Methods in Natural Language Processing (EMNLP)},
  year      = {2020},
  pages     = {6769--6781},
  publisher = {Association for Computational Linguistics},
  doi       = {10.18653/v1/2020.emnlp-main.550},
  url       = {https://aclanthology.org/2020.emnlp-main.550/}
}

@inproceedings{li2023compressing,
  author    = {Yucheng Li and Bo Dong and Frank Guerin and Chenghua Lin},
  title     = {Compressing context to enhance inference efficiency of large language models},
  booktitle = {Proceedings of the 2023 Conference on Empirical Methods in Natural Language Processing (EMNLP)},
  year      = {2023},
  pages     = {6342--6353},
  publisher = {Association for Computational Linguistics},
  doi       = {10.18653/v1/2023.emnlp-main.391},
  url       = {https://aclanthology.org/2023.emnlp-main.391/}
}

@inproceedings{pan2024llmlingua2,
  author    = {Zhuoshi Pan and Qianhui Wu and Huiqiang Jiang and Menglin Xia and Xufang Luo and Jue Zhang and Qingwei Lin and Victor R{\"u}hle and Yuqing Yang and Chin-Yew Lin and H. Vicky Zhao and Lili Qiu and Dongmei Zhang},
  title     = {{LLMLingua-2}: Data distillation for efficient and faithful task-agnostic prompt compression},
  booktitle = {Findings of the Association for Computational Linguistics: ACL 2024},
  year      = {2024},
  pages     = {963--981},
  publisher = {Association for Computational Linguistics},
  doi       = {10.18653/v1/2024.findings-acl.57},
  url       = {https://aclanthology.org/2024.findings-acl.57/}
}

@article{robertson2009bm25,
  author  = {Stephen Robertson and Hugo Zaragoza},
  title   = {The probabilistic relevance framework: {BM25} and beyond},
  journal = {Foundations and Trends in Information Retrieval},
  volume  = {3},
  number  = {4},
  pages   = {333--389},
  year    = {2009},
  doi     = {10.1561/1500000019},
  url     = {https://www.nowpublishers.com/article/DownloadEBook/INR-019}
}

@incollection{shapley1953value,
  author    = {Lloyd S. Shapley},
  title     = {A value for $n$-person games},
  booktitle = {Contributions to the Theory of Games, Volume II},
  editor    = {Harold W. Kuhn and Albert W. Tucker},
  pages     = {307--317},
  publisher = {Princeton University Press},
  year      = {1953},
  doi       = {10.1515/9781400881970-018}
}

@article{trivedi2022musique,
  author  = {Harsh Trivedi and Niranjan Balasubramanian and Tushar Khot and Ashish Sabharwal},
  title   = {{MuSiQue}: Multihop questions via single-hop question composition},
  journal = {Transactions of the Association for Computational Linguistics (TACL)},
  volume  = {10},
  pages   = {539--554},
  year    = {2022},
  doi     = {10.1162/tacl_a_00475},
  url     = {https://aclanthology.org/2022.tacl-1.31/}
}

@inproceedings{wagstaff2019limitations,
  author    = {Edward Wagstaff and Fabian Fuchs and Martin Engelcke and Ingmar Posner and Michael A. Osborne},
  title     = {On the limitations of representing functions on sets},
  booktitle = {Proceedings of the 36th International Conference on Machine Learning},
  year      = {2019},
  pages     = {6487--6494},
  volume    = {97},
  series    = {Proceedings of Machine Learning Research},
  publisher = {PMLR},
  url       = {https://proceedings.mlr.press/v97/wagstaff19a.html}
}

@inproceedings{yang2018hotpotqa,
  author    = {Zhilin Yang and Peng Qi and Saizheng Zhang and Yoshua Bengio and William Cohen and Ruslan Salakhutdinov and Christopher D. Manning},
  title     = {{HotpotQA}: A dataset for diverse, explainable multi-hop question answering},
  booktitle = {Proceedings of the 2018 Conference on Empirical Methods in Natural Language Processing (EMNLP)},
  year      = {2018},
  pages     = {2369--2380},
  publisher = {Association for Computational Linguistics},
  doi       = {10.18653/v1/D18-1259},
  url       = {https://aclanthology.org/D18-1259/}
}

@inproceedings{zaheer2018deepsets,
  title         = {Deep Sets},
  author        = {Manzil Zaheer and Satwik Kottur and Siamak Ravanbakhsh and Barnabas Poczos and Ruslan Salakhutdinov and Alexander J. Smola},
  booktitle     = {Advances in Neural Information Processing Systems},
  volume        = {30},
  year          = {2017},
  url           = {https://proceedings.neurips.cc/paper/2017/hash/f22e4747da1aa27e363d86d40ff442fe-Abstract.html}
}

@misc{nogueira2020passagererankingbert,
  title         = {Passage Re-ranking with {BERT}},
  author        = {Rodrigo Nogueira and Kyunghyun Cho},
  year          = {2019},
  eprint        = {1901.04085},
  archiveprefix = {arXiv},
  primaryclass  = {cs.IR},
  url           = {https://arxiv.org/abs/1901.04085}
}

@inproceedings{wang2020minilmdeepselfattentiondistillation,
  title         = {{MiniLM}: Deep Self-Attention Distillation for Task-Agnostic Compression of Pre-Trained Transformers},
  author        = {Wenhui Wang and Furu Wei and Li Dong and Hangbo Bao and Nan Yang and Ming Zhou},
  year          = {2020},
  booktitle     = {Advances in Neural Information Processing Systems},
  volume        = {33},
  pages         = {5776--5788},
  url           = {https://proceedings.neurips.cc/paper/2020/hash/3f5ee243547dee91fbd053c1c4a845aa-Abstract.html}
}

@inproceedings{reimers2019sentencebertsentenceembeddingsusing,
  title         = {Sentence-{BERT}: Sentence Embeddings using Siamese {BERT}-Networks},
  author        = {Nils Reimers and Iryna Gurevych},
  year          = {2019},
  booktitle     = {Proceedings of the 2019 Conference on Empirical Methods in Natural Language Processing and the 9th International Joint Conference on Natural Language Processing (EMNLP-IJCNLP)},
  pages         = {3982--3992},
  publisher     = {Association for Computational Linguistics},
  doi           = {10.18653/v1/D19-1410},
  url           = {https://aclanthology.org/D19-1410/}
}

@misc{do2026loocompleverageleaveoneoutstrategy,
  title         = {{LooComp}: Leverage Leave-One-Out Strategy to Encoder-only Transformer for Efficient Query-aware Context Compression},
  author        = {Thao Do and Dinh Phu Tran and An Vo and Seon Kwon Kim and Daeyoung Kim},
  year          = {2026},
  eprint        = {2603.09222},
  archiveprefix = {arXiv},
  primaryclass  = {cs.CL},
  url           = {https://arxiv.org/abs/2603.09222}
}

@inproceedings{xu2023recompimprovingretrievalaugmentedlms,
  title         = {{RECOMP}: Improving Retrieval-Augmented {LMs} with Context Compression and Selective Augmentation},
  author        = {Fangyuan Xu and Weijia Shi and Eunsol Choi},
  year          = {2024},
  booktitle     = {The Twelfth International Conference on Learning Representations},
  url           = {https://proceedings.iclr.cc/paper_files/paper/2024/hash/bda88ed2892f5e61c9a9bf215c566913-Abstract-Conference.html}
}

@inproceedings{lewis2021retrievalaugmentedgenerationknowledgeintensivenlp,
  title         = {Retrieval-Augmented Generation for Knowledge-Intensive {NLP} Tasks},
  author        = {Patrick Lewis and Ethan Perez and Aleksandra Piktus and Fabio Petroni and Vladimir Karpukhin and Naman Goyal and Heinrich K{\"u}ttler and Mike Lewis and Wen-tau Yih and Tim Rockt{\"a}schel and Sebastian Riedel and Douwe Kiela},
  year          = {2020},
  booktitle     = {Advances in Neural Information Processing Systems},
  volume        = {33},
  url           = {https://proceedings.neurips.cc/paper/2020/hash/6b493230205f780e1bc26945df7481e5-Abstract.html}
}

@misc{yuichi-tateno-2025-open-provence,
  url    = {https://github.com/hotchpotch/open_provence},
  title  = {{OpenProvence}: An Open-Source Implementation of Efficient and Robust Context Pruning for Retrieval-Augmented Generation},
  author = {Yuichi Tateno},
  year   = {2025}
}

@inproceedings{modarressi2025nolima,
  title     = {{N}o{L}i{M}a: Long-Context Evaluation Beyond Literal Matching},
  author    = {Modarressi, Ali and Deilamsalehy, Hanieh and Dernoncourt, Franck and Bui, Trung and Rossi, Ryan A. and Yoon, Seunghyun and Schuetze, Hinrich},
  booktitle = {Proceedings of the 42nd International Conference on Machine Learning},
  pages     = {44554--44570},
  year      = {2025},
  volume    = {267},
  series    = {Proceedings of Machine Learning Research},
  publisher = {PMLR},
  url       = {https://proceedings.mlr.press/v267/modarressi25a.html}
}

@inproceedings{taguchi2025adaptivek,
  title     = {Efficient Context Selection for Long-Context {QA}: No Tuning, No Iteration, Just Adaptive-$k$},
  author    = {Taguchi, Chihiro and Maekawa, Seiji and Bhutani, Nikita},
  booktitle = {Proceedings of the 2025 Conference on Empirical Methods in Natural Language Processing},
  pages     = {20105--20130},
  year      = {2025},
  publisher = {Association for Computational Linguistics},
  doi       = {10.18653/v1/2025.emnlp-main.1017},
  url       = {https://aclanthology.org/2025.emnlp-main.1017/}
}

@misc{meng2025dps,
  title         = {From Ranking to Selection: A Simple but Efficient Dynamic Passage Selector for Retrieval Augmented Generation},
  author        = {Meng, Siyuan and Liu, Junming and Chen, Yirong and Mao, Song and Cai, Pinlong and Yan, Guohang and Shi, Botian and Wang, Ding},
  year          = {2025},
  eprint        = {2508.09497},
  archiveprefix = {arXiv},
  primaryclass  = {cs.CL},
  url           = {https://arxiv.org/abs/2508.09497}
}

@inproceedings{deng2025influence,
  title     = {Influence Guided Context Selection for Effective Retrieval-Augmented Generation},
  author    = {Deng, Jiale and Shen, Yanyan and Pei, Ziyuan and Chen, Youmin and Huang, Linpeng},
  booktitle = {Advances in Neural Information Processing Systems},
  volume    = {38},
  year      = {2025},
  url       = {https://proceedings.neurips.cc/paper_files/paper/2025/hash/2a07497c94cd24b20faa3fd14d847037-Abstract-Conference.html}
}

@inproceedings{liu2025attribot,
  title     = {{AttriBoT}: A Bag of Tricks for Efficiently Approximating Leave-One-Out Context Attribution},
  author    = {Liu, Fengyuan and Kandpal, Nikhil and Raffel, Colin},
  booktitle = {The Thirteenth International Conference on Learning Representations},
  year      = {2025},
  url       = {https://openreview.net/forum?id=9kJperA2a4}
}

@misc{cohenwang2025attributionattention,
  title         = {Learning to Attribute with Attention},
  author        = {Cohen-Wang, Benjamin and Chuang, Yung-Sung and Madry, Aleksander},
  year          = {2025},
  eprint        = {2504.13752},
  archiveprefix = {arXiv},
  primaryclass  = {cs.LG},
  url           = {https://arxiv.org/abs/2504.13752}
}

@inproceedings{hwang2025exit,
  title     = {{EXIT}: Context-Aware Extractive Compression for Enhancing Retrieval-Augmented Generation},
  author    = {Taeho Hwang and Sukmin Cho and Soyeong Jeong and Hoyun Song and SeungYoon Han and Jong C. Park},
  booktitle = {Findings of the Association for Computational Linguistics: ACL 2025},
  pages     = {4895--4924},
  year      = {2025},
  publisher = {Association for Computational Linguistics},
  doi       = {10.18653/v1/2025.findings-acl.253},
  url       = {https://aclanthology.org/2025.findings-acl.253/}
}

@inproceedings{huang2025gumbel,
  title     = {Gumbel Reranking: Differentiable End-to-End Reranker Optimization},
  author    = {Siyuan Huang and Zhiyuan Ma and Jintao Du and Changhua Meng and Weiqiang Wang and Jingwen Leng and Minyi Guo and Zhouhan Lin},
  booktitle = {Proceedings of the 63rd Annual Meeting of the Association for Computational Linguistics (Volume 1: Long Papers)},
  pages     = {7142--7161},
  year      = {2025},
  publisher = {Association for Computational Linguistics},
  doi       = {10.18653/v1/2025.acl-long.354},
  url       = {https://aclanthology.org/2025.acl-long.354/}
}

@inproceedings{zerveas2022coder,
  title     = {{CODER}: An efficient framework for improving retrieval through {COntextual Document Embedding Reranking}},
  author    = {George Zerveas and Navid Rekabsaz and Daniel Cohen and Carsten Eickhoff},
  booktitle = {Proceedings of the 2022 Conference on Empirical Methods in Natural Language Processing},
  pages     = {10626--10644},
  year      = {2022},
  publisher = {Association for Computational Linguistics},
  doi       = {10.18653/v1/2022.emnlp-main.727},
  url       = {https://aclanthology.org/2022.emnlp-main.727/}
}

@inproceedings{song2024cmc,
  title     = {Comparing Neighbors Together Makes it Easy: Jointly Comparing Multiple Candidates for Efficient and Effective Retrieval},
  author    = {Jonghyun Song and Cheyon Jin and Wenlong Zhao and Andrew McCallum and Jay-Yoon Lee},
  booktitle = {Proceedings of the 2024 Conference on Empirical Methods in Natural Language Processing},
  pages     = {22255--22269},
  year      = {2024},
  publisher = {Association for Computational Linguistics},
  doi       = {10.18653/v1/2024.emnlp-main.1242},
  url       = {https://aclanthology.org/2024.emnlp-main.1242/}
}

@inproceedings{zhang2021muppets,
  title     = {Learning to Rank in the Age of {Muppets}: Effectiveness--Efficiency Tradeoffs in Multi-Stage Ranking},
  author    = {Yue Zhang and Chengcheng Hu and Yuqi Liu and Hui Fang and Jimmy Lin},
  booktitle = {Proceedings of the Second Workshop on Simple and Efficient Natural Language Processing},
  pages     = {64--73},
  year      = {2021},
  publisher = {Association for Computational Linguistics},
  doi       = {10.18653/v1/2021.sustainlp-1.8},
  url       = {https://aclanthology.org/2021.sustainlp-1.8/}
}

@inproceedings{zhao2025funnelrag,
  title     = {{FunnelRAG}: A Coarse-to-Fine Progressive Retrieval Paradigm for {RAG}},
  author    = {Xinping Zhao and Yan Zhong and Zetian Sun and Xinshuo Hu and Zhenyu Liu and Dongfang Li and Baotian Hu and Min Zhang},
  booktitle = {Findings of the Association for Computational Linguistics: NAACL 2025},
  pages     = {3029--3046},
  year      = {2025},
  publisher = {Association for Computational Linguistics},
  doi       = {10.18653/v1/2025.findings-naacl.165},
  url       = {https://aclanthology.org/2025.findings-naacl.165/}
}

@inproceedings{laitenberger2025stronger,
  title     = {Stronger Baselines for Retrieval-Augmented Generation with Long-Context Language Models},
  author    = {Alex Laitenberger and Christopher D. Manning and Nelson F. Liu},
  booktitle = {Proceedings of the 2025 Conference on Empirical Methods in Natural Language Processing},
  pages     = {32559--32569},
  year      = {2025},
  publisher = {Association for Computational Linguistics},
  doi       = {10.18653/v1/2025.emnlp-main.1656},
  url       = {https://aclanthology.org/2025.emnlp-main.1656/}
}

@inproceedings{yuan2026ebcar,
  title     = {Embedding-Based Context-Aware Reranker},
  author    = {Ye Yuan and Mohammad Amin Shabani and Siqi Liu},
  booktitle = {The Fourteenth International Conference on Learning Representations},
  year      = {2026},
  url       = {https://openreview.net/forum?id=OBMcxeSK5U}
}

@inproceedings{sun2025dynamicrag,
  title     = {{DynamicRAG}: Leveraging Outputs of Large Language Model as Feedback for Dynamic Reranking in Retrieval-Augmented Generation},
  author    = {Jiashuo Sun and Xianrui Zhong and Sizhe Zhou and Jiawei Han},
  booktitle = {Advances in Neural Information Processing Systems},
  volume    = {38},
  year      = {2025},
  url       = {https://proceedings.neurips.cc/paper_files/paper/2025/hash/f69d2246a85e28c3e63dee3daf8d4af8-Abstract-Conference.html}
}

@misc{sun2026barrag,
  title         = {Rethinking the Reranker: Boundary-Aware Evidence Selection for Robust Retrieval-Augmented Generation},
  author        = {Jiashuo Sun and Pengcheng Jiang and Saizhuo Wang and Jiajun Fan and Heng Wang and Siru Ouyang and Ming Zhong and Yizhu Jiao and Chengsong Huang and Xueqiang Xu and Pengrui Han and Peiran Li and Jiaxin Huang and Ge Liu and Heng Ji and Jiawei Han},
  year          = {2026},
  eprint        = {2602.03689},
  archiveprefix = {arXiv},
  primaryclass  = {cs.CL},
  url           = {https://arxiv.org/abs/2602.03689}
}

@misc{chen2026shapleycontextpruningcooperative,
  title         = {{Shapley} Context Pruning: A Cooperative Game Perspective for Context Reranking and Pruning},
  author        = {Yanqiao Chen and Dongsheng Hou and Yuhan Rui and Zhen Cao and Yepang Liu},
  year          = {2026},
  eprint        = {2607.16209},
  archiveprefix = {arXiv},
  primaryclass  = {cs.AI},
  url           = {https://arxiv.org/abs/2607.16209}
}

@inproceedings{santhanam2022colbertv2,
  title     = {{ColBERTv2}: Effective and Efficient Retrieval via Lightweight Late Interaction},
  author    = {Keshav Santhanam and Omar Khattab and Jon Saad-Falcon and Christopher Potts and Matei Zaharia},
  booktitle = {Proceedings of the 2022 Conference of the North American Chapter of the Association for Computational Linguistics: Human Language Technologies},
  pages     = {3715--3734},
  year      = {2022},
  publisher = {Association for Computational Linguistics},
  doi       = {10.18653/v1/2022.naacl-main.272},
  url       = {https://aclanthology.org/2022.naacl-main.272/}
}

@inproceedings{es2024ragas,
  title     = {{RAGAs}: Automated Evaluation of Retrieval Augmented Generation},
  author    = {Es, Shahul and James, Jithin and Espinosa Anke, Luis and Schockaert, Steven},
  booktitle = {Proceedings of the 18th Conference of the European Chapter of the Association for Computational Linguistics: System Demonstrations},
  pages     = {150--158},
  year      = {2024},
  publisher = {Association for Computational Linguistics},
  doi       = {10.18653/v1/2024.eacl-demo.16},
  url       = {https://aclanthology.org/2024.eacl-demo.16/}
}

@misc{papadimitriou2024ragplayground,
  title         = {{RAG Playground}: A Framework for Systematic Evaluation of Retrieval Strategies and Prompt Engineering in {RAG} Systems},
  author        = {Ioannis Papadimitriou and Ilias Gialampoukidis and Stefanos Vrochidis and Ioannis Kompatsiaris},
  year          = {2024},
  eprint        = {2412.12322},
  archiveprefix = {arXiv},
  primaryclass  = {cs.LG},
  url           = {https://arxiv.org/abs/2412.12322}
}

@inproceedings{murugaraj2026ragvue,
  title     = {{RAGVUE}: A Diagnostic View for Explainable and Automated Evaluation of Retrieval-Augmented Generation},
  author    = {Keerthana Murugaraj and Salima Lamsiyah and Martin Theobald},
  booktitle = {Proceedings of the 19th Conference of the European Chapter of the Association for Computational Linguistics (Volume 3: System Demonstrations)},
  pages     = {512--526},
  year      = {2026},
  publisher = {Association for Computational Linguistics},
  doi       = {10.18653/v1/2026.eacl-demo.35},
  url       = {https://aclanthology.org/2026.eacl-demo.35/}
}

@misc{rengo2025scarf,
  title         = {A System for Comprehensive Assessment of {RAG} Frameworks},
  author        = {Mattia Rengo and Senad Beadini and Domenico Alfano and Roberto Abbruzzese},
  year          = {2025},
  eprint        = {2504.07803},
  archiveprefix = {arXiv},
  primaryclass  = {cs.CL},
  url           = {https://arxiv.org/abs/2504.07803}
}

@misc{simon2024ragevaluation,
  title         = {A Methodology for Evaluating {RAG} Systems: A Case Study on Configuration Dependency Validation},
  author        = {Sebastian Simon and Alina Mailach and Johannes Dorn and Norbert Siegmund},
  year          = {2024},
  eprint        = {2410.08801},
  archiveprefix = {arXiv},
  primaryclass  = {cs.SE},
  url           = {https://arxiv.org/abs/2410.08801}
}

@inproceedings{cormack2009rrf,
  title     = {Reciprocal Rank Fusion outperforms {Condorcet} and individual Rank Learning Methods},
  author    = {Gordon V. Cormack and Charles L. A. Clarke and Stefan B{\"u}ttcher},
  booktitle = {Proceedings of the 32nd International ACM SIGIR Conference on Research and Development in Information Retrieval},
  pages     = {758--759},
  year      = {2009},
  publisher = {Association for Computing Machinery},
  url       = {https://plg.uwaterloo.ca/~gvcormac/cormacksigir09-rrf.pdf}
}

@misc{gemmateam2025gemma3,
  title         = {{Gemma 3} Technical Report},
  author        = {{Gemma Team} and Aishwarya Kamath and Johan Ferret and others},
  year          = {2025},
  eprint        = {2503.19786},
  archiveprefix = {arXiv},
  primaryclass  = {cs.CL},
  url           = {https://arxiv.org/abs/2503.19786}
}

@misc{grattafiori2024llama3,
  title         = {The {Llama 3} Herd of Models},
  author        = {Aaron Grattafiori and Abhimanyu Dubey and Abhinav Jauhri and others},
  year          = {2024},
  eprint        = {2407.21783},
  archiveprefix = {arXiv},
  primaryclass  = {cs.AI},
  url           = {https://arxiv.org/abs/2407.21783v3}
}

@misc{baai2023bge,
  title        = {{bge-small-en-v1.5} Model Card},
  author       = {{Beijing Academy of Artificial Intelligence}},
  year         = {2023},
  howpublished = {Hugging Face},
  url          = {https://huggingface.co/BAAI/bge-small-en-v1.5},
  note         = {Accessed: 2026-09-24}
}
\bibliographystyle{iclr2027/iclr2027_conference}

\clearpage
\appendix

\section{Experiment Results and Reproducibility Details}\label{app:exp_results}

\subsection{Datasets, Models, and Training Metadata}

\begin{table}[H]
\centering
\small
\caption{Dataset roles and evaluation units.}
\label{tab:dataset_setup_new}
\resizebox{\textwidth}{!}{%
\begin{tabular}{lcll}
\toprule
\textbf{Dataset} & \textbf{Examples} & \textbf{Role} & \textbf{Primary metrics} \\
\midrule
MS MARCO v2.1 & 452,644 / 25,147 / 25,147 & Passage-level train / validation / held-out ranking & NPRS, MAP, nDCG, recall \\
HotpotQA & 612 & Zero-shot sentence ranking & NPRS, MAP, nDCG, recall \\
2WikiMultiHopQA & 7,737 / 600 & Zero-shot sentence ranking / cascade QA & Ranking, evidence recall, EM/F1 \\
MuSiQue & 997 & Zero-shot paragraph ranking & NPRS, MAP, nDCG, recall \\
NoLiMa & 870 per method/reader & 32K sentence-level direct and cascade evaluation & Exact retention, answer accuracy, latency \\
\bottomrule
\end{tabular}
}
\end{table}

\begin{table}[H]
\centering
\small
\caption{Cacheable model and execution metadata.}
\label{tab:model_setup_new}
\resizebox{\textwidth}{!}{%
\begin{tabular}{ll}
\toprule
\textbf{Component} & \textbf{Setting} \\
\midrule
Frozen encoder & MiniLM-L6-v2, 384-dimensional normalized candidate embeddings \\
Late-interaction endpoint & ColBERTv2 MS MARCO checkpoint, 128-dimensional token matrices, exact fixed-pool MaxSim \\
Feature schema & $[q,s,q\odot s,|q-s|,\cos(q,s)]$, normalized embeddings, 1,537 dimensions \\
Learned scorer & $\psi$: 1537--896--896--448; $\rho$: 448--896--448--1; LayerNorm, ReLU, dropout 0.1; 3,394,497 parameters \\
Training & seed 42, 10 epochs, learning rate $10^{-3}$, weight decay $10^{-4}$, margin 0.15 \\
Shapley objective & 10 sampled permutations per training context; singleton validation/deployment \\
Pointwise objective & singleton credit training; singleton validation \\
Readers & \texttt{unsloth/gemma-3-12b-it}; \texttt{meta-llama/Llama-3.1-8B-Instruct}; bfloat16, greedy decoding \\
Timing & A800 GPU, isolated processes, explicit CUDA synchronization; 50 warm-ups for the unified CE/Provence panel, 3 for ColBERTv2 \\
\bottomrule
\end{tabular}
}
\end{table}

Validation NPRS selects epoch 5 for Cache-Shapley and epoch 4 for Cache-Pointwise. Both models use the same split, encoder, feature schema, optimization budget, and validation rule. Primary results use seed 42; robustness runs repeat both objectives with seeds 123 and 2026. The BGE-small-en-v1.5 control \citep{baai2023bge} retains singleton pointwise training and tests the learned fusion with a stronger frozen representation.

\paragraph{Reranker implementations.} CrossEncoder uses \nolinkurl{cross-encoder/ms-marco-MiniLM-L-6-v2}. Provence denotes the OpenProvence checkpoint \nolinkurl{hotchpotch/open-provence-reranker-v1-gte-modernbert-base} \citep{yuichi-tateno-2025-open-provence}. CrossEncoder scores query--sentence pairs independently. OpenProvence receives the entire admitted, pre-split context and returns aligned sentence scores. Within a cascade this context follows coarse-score order; the final reader input returns to source order. ColBERTv2 uses \nolinkurl{colbert-ir/colbertv2.0}, exact MaxSim, and CPU-cached fp16 document-token matrices. The single-vector cache stores normalized float32 sentence embeddings on CPU.

\paragraph{MS MARCO training-context construction.} The loader reads the Hugging Face \texttt{microsoft/ms\_marco} v2.1 training split in release order and retains every nonempty \texttt{passage\_text} longer than ten stripped characters as one candidate $c_i$. Its released \texttt{is\_selected} flag supplies $y_i\in\{0,1\}$. All retained $y_i{=}1$ passages form $\mathcal P$, and all retained $y_i{=}0$ passages form $\mathcal D$. Training negatives consist exclusively of released within-context $y_i{=}0$ passages. Passage granularity follows the release's passage-level annotation; sentence-level propagation would introduce unverified supervision. We keep contexts with at least one positive and 2--50 retained candidates. The pairwise loss averages all within-context pairs in $\mathcal P\times\mathcal D$; a context with an empty side contributes zero loss. Filtering yields 502,938 contexts, deterministically sliced in release order into 452,644/25,147/25,147 training/validation/held-out contexts. The unchanged scorer is then evaluated in zero-shot mode on dataset-native units, yielding passage-to-sentence transfer on HotpotQA, 2WikiMultiHopQA, NoLiMa, and the sentence-level cascades, and passage-to-paragraph transfer on MuSiQue.

\paragraph{Evaluation protocols and statistical units.} Ranking uses the complete 25,147-query MS MARCO held-out split and the fixed HotpotQA, 2WikiMultiHopQA, and MuSiQue splits in Table~\ref{tab:dataset_setup_new}. NoLiMa crosses all 58 official tests with five book haystacks and three insertion depths at 32K, yielding 870 matched outcomes per method and reader. Direct selection and every final cascade output obey the same 30\% token cap. Extractive methods reuse fixed selections across readers and reproduce identical retention, so only the reader changes in their cross-reader comparisons. The primary cascades admit 50\% of sentences to CrossEncoder or Provence; the auxiliary CrossEncoder sweep compares 50\% and 70\%. The ColBERTv2 extension uses the released MS MARCO checkpoint and exact MaxSim over the fixed sentence pool. Its two-stage panel admits 50\% or 70\% to ColBERTv2; its three-stage panel admits 70\% to ColBERTv2 and 50\% to CrossEncoder or Provence. This evaluation covers fixed-pool reranking and does not include corpus-scale PLAID indexing. The 2Wiki experiment fixes 600 validation questions, admits 70\% of the original sentences to the precise stage, and retains 50\% of the original sentence count after that stage.

Full-ranking metrics are macro-averaged over contexts. NPRS is $1-\frac{1}{n|\mathcal P|}\sum_{i\in\mathcal P}\operatorname{rank}(i)$ for zero-indexed descending ranks; MAP, nDCG, nDCG@10, and evidence recall provide standard ranking views. Reader metrics are computed after restoring selected units to source order. Paired intervals resample queries. NoLiMa resamples 290 haystack--test clusters and keeps all three depths of a cluster together. Timing runs each method in an isolated process on the same A800 GPU model, uses the warm-up counts in Table~\ref{tab:model_setup_new}, explicitly synchronizes CUDA, and includes query encoding, cache-miss candidate encoding, coarse scoring, Top-$K$, precise reranking, budget selection, and context materialization. Model loading and shared input parsing precede the timed region. Candidate embeddings are cached lazily; aggregate timing includes initial corpus-indexing requests, while warm-cache statistics exclude cache misses. Speedups divide mean latencies; p50 and p95 are reported separately.
\vspace{-2pt}
\begin{table}[H]
\centering
\small
\caption{QA prompt design and decoding settings. Max is the maximum number of generated tokens.}
\label{tab:prompt_usage_new}
\begin{tabularx}{\linewidth}{C{2.6cm}Ycc}
\toprule
\textbf{Evaluation} & \multicolumn{1}{c}{\textbf{Prompt instruction}} & \textbf{Max} & \textbf{Temp.} \\
\midrule
2WikiMH, Gemma & Answer from the retained context with a concise 1--5 word response; return only ``yes'' or ``no'' for binary questions. & 64 & 0.0 \\
NoLiMa, Gemma & Follow the official test prompt and return a concise answer used by answer-containment scoring. & 192 & 0.0 \\
NoLiMa, Llama & The same official test prompt and answer-containment scoring used for Gemma. & 192 & 0.0 \\
\bottomrule
\end{tabularx}
\end{table}
\vspace{-2pt}
These settings define reproducible comparisons with explicit data, model, budget, decoding, and timing choices.
\vspace{-5pt}
\subsection{Algorithmic Procedures}\label{sec:algorithms_new}

The procedures below separate three reusable operations: context-unit credit computation, matched cacheable training under a specified credit objective, and deployment in a coarse-to-precise cascade.

Training-credit computation is the only objective-specific step; both learned models enter the same cacheable cascade with singleton scores.

\begingroup
\setlength{\intextsep}{7pt}

\begin{algorithm}[H]
\small
\caption{Shapley or Singleton Credit Readout}
\label{alg:credit_readout_new}
\begin{algorithmic}[1]
\Require Value function $v_\theta$, candidates $C=\{c_i\}_{i=1}^n$, permutations $M$, mode $r$
\If{$r=\mathrm{Singleton}$}
    \State \Return $[v_\theta(\{c_1\}),\ldots,v_\theta(\{c_n\})]$
\EndIf
\State $\phi_i\gets 0$ for every $i$
\For{$m=1,\ldots,M$}
    \State sample a uniform permutation $\pi$ of $\{1,\ldots,n\}$; $S\gets\emptyset$; $u\gets v_\theta(\emptyset)$
    \For{$i$ in the order $\pi$}
        \State $u'\gets v_\theta(S\cup\{c_i\})$; $\phi_i\gets\phi_i+(u'-u)/M$
        \State $S\gets S\cup\{c_i\}$; $u\gets u'$
    \EndFor
\EndFor
\State \Return $[\phi_1,\ldots,\phi_n]$
\end{algorithmic}
\end{algorithm}

\begin{algorithm}[H]
\small
\caption{Matched Pointwise or Shapley Credit Training}
\label{alg:matched_training_new}
\begin{algorithmic}[1]
\Require Training contexts $(q,C,\mathcal{P},\mathcal{D})$, frozen encoder $E$, objective $o$, margin $\epsilon$
\For{each minibatch}
    \For{each context $(q,C,\mathcal{P},\mathcal{D})$}
        \State encode $q$ once and retrieve cached candidate vectors; construct pair features $x_i$
        \State build the set value $v_\theta(S)=\rho(\sum_{i\in S}\psi(x_i))$
        \If{$o=\mathrm{Pointwise}$}
            \State $r_i\gets v_\theta(\{c_i\})$ for every $i$
        \Else
            \State $r\gets\Call{CreditReadout}{v_\theta,C,M,\mathrm{Shapley}}$
        \EndIf
        \If{$|\mathcal P||\mathcal D|>0$}
            \State $\mathcal{L}\gets\frac{1}{|\mathcal{P}||\mathcal{D}|}\sum_{i\in\mathcal{P}}\sum_{j\in\mathcal{D}}\max\{0,\epsilon-r_i+r_j\}$
        \Else
            \State $\mathcal L\gets 0$
        \EndIf
    \EndFor
    \State update $\theta$ with the minibatch mean loss
\EndFor
\State \Return validation-selected parameters $\theta$
\end{algorithmic}
\end{algorithm}

\begin{algorithm}[H]
\small
\caption{Cacheable Semantic Cascade Inference}
\label{alg:semantic_cascade_new}
\begin{algorithmic}[1]
\Require Query $q$, candidates $C$, cached vectors $\{z_i\}$, coarse model $v_\theta$, precise reranker $h$, coarse fraction $\alpha$, final budget $B$
\State encode $q$ once; construct $x_i=[z_q;z_i;z_q\odot z_i;|z_q-z_i|;\cos(z_q,z_i)]$
\State $s_i\gets v_\theta(\{c_i\})$ for every candidate
\State $C_1\gets\operatorname{Top}_{\min(n,\max(1,\operatorname{round}(\alpha n)))}(C;s)$
\State evaluate precise scores $t\gets h(q,C_1)$ for the admitted candidates
\State $C_2\gets\operatorname{BudgetTop}_{B}(C_1;t)$
\State restore the source order of $C_2$ and \Return $C_2$
\end{algorithmic}
\end{algorithm}

\endgroup
\vspace{-19pt}

\subsection{Standard Ranking Metrics}\label{sec:cache_metrics_supp}

For contexts containing only positive or only negative labels, the evaluator assigns NPRS 0.5. Ranking recall at fraction $r$ uses $\max(1,\lfloor rn\rfloor)$ candidates; the sentence-count reader budgets use rounding as specified in Section~\ref{sec:method}.

\begingroup
\setlength{\intextsep}{7pt}

\begin{table}[H]
\centering
\small
\setlength{\tabcolsep}{8pt}
\caption{Held-out MS MARCO metrics for the cacheable models and frozen MiniLM cosine.}
\label{tab:cache_standard_metrics}
\begin{tabular}{lcccc}
\toprule
\textbf{Method} & \textbf{NPRS} & \textbf{MAP} & \textbf{nDCG@10} & \textbf{R@30\%} \\
\midrule
Frozen MiniLM cosine & 0.8352 & 0.6085 & 0.7058 & 0.7448 \\
Cache-Pointwise & \textbf{0.8479} & 0.6303 & 0.7225 & \textbf{0.7689} \\
Cache-Shapley & 0.8477 & \textbf{0.6314} & \textbf{0.7233} & 0.7681 \\
\bottomrule
\end{tabular}
\end{table}

\begin{table}[H]
\centering
\small
\setlength{\tabcolsep}{8pt}
\caption{Zero-shot ranking metrics for cacheable models. Values use fixed dataset splits and no target-dataset tuning.}
\label{tab:cache_zeroshot_full}
\begin{tabular}{llcccc}
\toprule
\textbf{Dataset} & \textbf{Method} & \textbf{NPRS} & \textbf{MAP} & \textbf{nDCG@10} & \textbf{R@30\%} \\
\midrule
\multirow{3}{*}{HotpotQA}
& MiniLM & 0.8109 & 0.5097 & 0.5960 & 0.7328 \\
& Cache-PW & 0.8321 & \textbf{0.5406} & \textbf{0.6336} & 0.7683 \\
& Cache-Sh. & \textbf{0.8327} & 0.5367 & 0.6296 & \textbf{0.7739} \\
\midrule
\multirow{3}{*}{2WikiMH}
& MiniLM & 0.8253 & 0.6134 & 0.7017 & 0.7462 \\
& Cache-PW & 0.8388 & 0.6331 & 0.7211 & 0.7622 \\
& Cache-Sh. & \textbf{0.8389} & \textbf{0.6335} & \textbf{0.7212} & \textbf{0.7637} \\
\midrule
\multirow{3}{*}{MuSiQue}
& MiniLM & \textbf{0.7478} & \textbf{0.5465} & \textbf{0.6449} & \textbf{0.6459} \\
& Cache-PW & 0.7305 & 0.5122 & 0.6092 & 0.6275 \\
& Cache-Sh. & 0.7285 & 0.5066 & 0.6050 & 0.6140 \\
\bottomrule
\end{tabular}
\end{table}

\endgroup
\vspace{-2pt}
Across the reported ranking metrics, both learned scorers improve on frozen MiniLM on MS MARCO, HotpotQA, and 2WikiMultiHopQA, while MuSiQue consistently favors frozen cosine.

\subsection{Stronger-Encoder Control}\label{sec:bge_control_supp}

\vspace{-2pt}

\begin{table}[H]
\centering
\small
\setlength{\tabcolsep}{8pt}
\caption{Seed-42 BGE-small-en-v1.5 control. The frozen cosine and learned pointwise head share the revision-locked encoder and query prefix. $\Delta$ is learned minus cosine. NoLiMa uses the same 32K inputs, 30\% token cap, and Gemma 3 12B reader as the main evaluation.}
\label{tab:bge_control}
\begin{tabular}{llccc}
\toprule
\textbf{Evaluation} & \textbf{Metric} & \textbf{BGE cosine} & \textbf{BGE-PW} & \textbf{$\Delta$} \\
\midrule
MS MARCO & NPRS & 0.8472 & \textbf{0.8598} & +0.0126 \\
HotpotQA & NPRS & 0.8621 & \textbf{0.8665} & +0.0044 \\
2WikiMH & NPRS & 0.8638 & \textbf{0.8731} & +0.0093 \\
MuSiQue & NPRS & \textbf{0.7859} & 0.7395 & $-0.0464$ \\
NoLiMa & Retention (\%) & 72.53 & \textbf{78.28} & +5.75 \\
NoLiMa & QA (\%) & \textbf{29.31} & 28.28 & $-1.03$ \\
\bottomrule
\end{tabular}
\end{table}
\vspace{-4pt}

The learned BGE head improves held-out MS MARCO NPRS by 0.0126 [0.0113, 0.0140] and 2WikiMultiHopQA by 0.0093 [0.0081, 0.0105]. Its HotpotQA NPRS difference is 0.0044 [$-0.0001$, 0.0090], and MuSiQue favors frozen cosine by 0.0464 [0.0399, 0.0529]. On NoLiMa, the learned head raises exact retention by 5.75 points and changes reader accuracy by $-1.03$ points. We do not evaluate a matched BGE Shapley model. This control shows that the learned head's benefit depends on the target task and that higher evidence retention need not improve reader accuracy.

\subsection{NoLiMa Direct References and Cascade Decomposition}\label{sec:cascade_decomp_supp}

\begingroup
\setlength{\intextsep}{4pt}

\begin{table}[H]
\centering
\small
\setlength{\tabcolsep}{7pt}
\caption{Direct NoLiMa 32K performance under a 30\% token cap. Retention is reader-independent; QA is answer accuracy over the same 870 outcomes with Gemma 3 12B (G) and Llama 3.1 8B (L). LLMLingua2 performs token-level compression, and QC-SCP from \citet{chen2026shapleycontextpruningcooperative} was evaluated only with Gemma.}
\label{tab:nolima_direct}
\begin{tabular}{lccc}
\toprule
\textbf{Method} & \textbf{Retention} & \textbf{QA-G} & \textbf{QA-L} \\
\midrule
Random & 28.51 & 13.68 & 11.38 \\
BM25 & 31.61 & 15.52 & 6.44 \\
LLMLingua2 & -- & 16.09 & 12.76 \\
CrossEncoder & 50.69 & 21.49 & 19.66 \\
Provence & 50.34 & 22.30 & 16.44 \\
Frozen MiniLM cosine & 72.18 & 27.70 & 21.49 \\
QC-SCP-Singleton & 58.05 & 27.59 & -- \\
Cache-Shapley & 76.44 & 30.34 & 20.69 \\
\textbf{Cache-Pointwise} & \textbf{86.09} & \textbf{31.49} & \textbf{26.55} \\
\bottomrule
\end{tabular}
\end{table}

\begin{table}[H]
\centering
\small
\setlength{\tabcolsep}{7pt}
\caption{Complete NoLiMa 32K cascade panel with Gemma 3 12B. Latency is synchronized p50 online pruning time; speedup divides the corresponding direct reranker's mean latency by the cascade's mean latency. Peak memory is the maximum per-request allocated GPU-memory peak. Retention and QA use the same 870 paired contexts.}
\label{tab:nolima_cascade}
\begin{tabular*}{\linewidth}{llccccc@{}}
\toprule
\textbf{Precise stage} & \textbf{Coarse stage} & \textbf{p50 ms} & \textbf{Speedup} & \textbf{Peak GiB} & \textbf{Retention} & \textbf{QA} \\
\midrule
\multirow{5}{*}{CrossEncoder}
& None & 237.0 & 1.00$\times$ & 0.27 & 0.5069 & 0.2149 \\
& BM25 & 152.1 & 1.55$\times$ & 0.27 & 0.4471 & 0.1931 \\
& MiniLM & 183.0 & 1.27$\times$ & 0.81 & 0.6655 & 0.2609 \\
& Cache-Shapley & 151.3 & 1.57$\times$ & 0.82 & 0.7057 & 0.2552 \\
& \textbf{Cache-Pointwise} & \textbf{149.6} & \textbf{1.58$\times$} & 0.82 & \textbf{0.7253} & \textbf{0.2851} \\
\midrule
\multirow{5}{*}{Provence}
& None & 755.0 & 1.00$\times$ & 32.17 & 0.5034 & 0.2230 \\
& BM25 & 482.0 & 1.68$\times$ & 23.74 & 0.4069 & 0.1966 \\
& MiniLM & 456.5 & 1.74$\times$ & \textbf{19.29} & 0.6897 & 0.2736 \\
& Cache-Shapley & \textbf{421.9} & \textbf{1.88$\times$} & 19.73 & 0.7023 & 0.2897 \\
& \textbf{Cache-Pointwise} & 428.5 & 1.86$\times$ & 19.89 & \textbf{0.7713} & \textbf{0.3172} \\
\bottomrule
\end{tabular*}
\end{table}

\endgroup

\begin{table}[H]
\centering
\small
\caption{Detailed NoLiMa cascade timing and evidence decomposition. Coarse retention is evidence survival after the first stage. Final-given-coarse is the conditional survival rate through the precise stage.}
\label{tab:cascade_decomposition}
\begin{tabular*}{\linewidth}{llccccc@{}}
\toprule
\textbf{Precise} & \textbf{Coarse} & \textbf{p95 ms} & \textbf{Coarse ret.} & \textbf{Final ret.} & \textbf{Final given coarse} & \textbf{Token load} \\
\midrule
\multirow{4}{*}{CrossEncoder}
& BM25 & 173.9 & 0.6230 & 0.4471 & 0.7177 & 0.585 \\
& MiniLM & 245.5 & 0.8632 & 0.6655 & 0.7710 & 0.493 \\
& Cache-PW & 166.2 & \textbf{0.9713} & \textbf{0.7253} & 0.7467 & 0.515 \\
& Cache-Sh. & 166.9 & 0.8874 & 0.7057 & \textbf{0.7953} & 0.504 \\
\midrule
\multirow{4}{*}{Provence}
& BM25 & 532.4 & 0.6230 & 0.4069 & 0.6531 & 0.585 \\
& MiniLM & 484.8 & 0.8632 & 0.6897 & \textbf{0.7989} & 0.493 \\
& Cache-PW & 448.1 & \textbf{0.9713} & \textbf{0.7713} & 0.7941 & 0.515 \\
& Cache-Sh. & \textbf{445.3} & 0.8874 & 0.7023 & 0.7915 & 0.504 \\
\bottomrule
\end{tabular*}
\end{table}

The five initial corpus-indexing requests are included in aggregate means and excluded by the warm-cache definition. Candidate encoding on cache misses is timed; model loading remains outside the online timing boundary. Direct CrossEncoder and Provence p95 latencies are 267.8 and 980.4 ms. Their cacheable pointwise cascades reduce p95 latency to 166.2 and 448.1 ms.

\begingroup
\setlength{\intextsep}{7.2pt}

\begin{table}[H]
\centering
\scriptsize
\setlength{\tabcolsep}{5pt}
\caption{Paired NoLiMa stage contrasts with Gemma. Entries are percentage-point difference [95\% cluster-bootstrap interval] over 870 matched contexts. Coarse contrasts compare with the precise-only system; precise contrasts compare with the corresponding coarse-only system.}
\label{tab:stage_signature_ci}
\begin{tabular*}{\linewidth}{llcccc@{}}
\toprule
$g$ & $h$ & $\Delta_g^{\rm Ret}$ & $\Delta_h^{\rm Ret}$ & $\Delta_g^{\rm QA}$ & $\Delta_h^{\rm QA}$ \\
\midrule
Cache-PW & CrossEncoder & +21.84 [18.39, 25.29] & $-13.56$ [$-17.13$, $-10.11$] & +7.01 [4.37, 9.77] & $-2.99$ [$-5.86$, $-0.11$] \\
Cache-Sh. & CrossEncoder & +19.89 [16.32, 23.68] & $-5.86$ [$-9.20$, $-2.53$] & +4.02 [1.15, 6.90] & $-4.83$ [$-7.70$, $-2.07$] \\
Cache-PW & Provence & +26.78 [22.53, 30.92] & $-8.97$ [$-12.18$, $-5.75$] & +9.43 [6.44, 12.41] & +0.23 [$-2.30$, 2.87] \\
Cache-Sh. & Provence & +19.89 [15.29, 24.25] & $-6.21$ [$-9.66$, $-2.87$] & +6.67 [3.79, 9.54] & $-1.38$ [$-3.80$, 1.15] \\
\bottomrule
\end{tabular*}
\end{table}

\begin{table}[H]
\centering
\small
\setlength{\tabcolsep}{6pt}
\caption{Complete NoLiMa knowledge-type answer accuracy with 95\% haystack--test cluster-bootstrap intervals, in percent. Ret. is exact evidence retention. Common and World contain 240 and 630 outcomes.}
\label{tab:knowledge_slices_ci}
\begin{tabular*}{\linewidth}{lcccc@{}}
\toprule
\textbf{Method} & \textbf{Common Acc. [95\% CI]} & \textbf{Common Ret.} & \textbf{World Acc. [95\% CI]} & \textbf{World Ret.} \\
\midrule
Random & 20.00 [14.58, 25.83] & 27.08 & 11.27 [8.73, 13.81] & 29.05 \\
BM25 & 37.50 [33.33, 41.67] & 49.17 & 7.14 [5.08, 9.52] & 24.92 \\
LLMLingua2 & 27.92 [21.67, 34.58] & -- & 11.59 [9.05, 14.29] & -- \\
CrossEncoder & 45.42 [37.50, 53.75] & 70.42 & 12.38 [9.52, 15.40] & 43.17 \\
Provence & 42.92 [35.42, 50.42] & 70.83 & 14.44 [11.59, 17.62] & 42.54 \\
Cache-Shapley & 51.25 [44.58, 57.92] & 95.00 & 22.38 [18.57, 26.19] & 69.37 \\
Cache-Pointwise & 49.17 [42.92, 55.83] & \textbf{95.42} & \textbf{24.76 [20.63, 28.89]} & \textbf{82.54} \\
Cache-Sh.$\rightarrow$CE & 46.67 [40.42, 52.92] & 85.42 & 17.46 [14.13, 20.95] & 64.92 \\
Cache-PW$\rightarrow$CE & 47.92 [40.42, 55.00] & 85.00 & 21.11 [17.46, 24.92] & 67.78 \\
Cache-Sh.$\rightarrow$Prov. & 54.17 [46.67, 61.67] & 88.33 & 19.37 [15.87, 23.02] & 63.33 \\
Cache-PW$\rightarrow$Prov. & \textbf{54.58 [47.92, 61.25]} & 93.75 & 23.02 [19.05, 26.98] & 70.79 \\
\bottomrule
\end{tabular*}
\end{table}

\begin{table}[H]
\centering
\scriptsize
\setlength{\tabcolsep}{6.5pt}
\caption{NoLiMa evidence-transition rates by knowledge type. CERR and CEBR use the corresponding precise-only reranker as reference; all entries are percent [95\% haystack--test cluster-bootstrap interval]. The net column is the resulting final-retention contrast in percentage points.}
\label{tab:evidence_repair_ci}
\begin{tabular*}{\linewidth}{lllcccc@{}}
\toprule
\textbf{Coarse} & \textbf{Precise} & \textbf{Slice} & \textbf{Miss/Hit $n$} & \textbf{CERR} & \textbf{CEBR} & \textbf{Net $\Delta_g^{\rm Ret}$} \\
\midrule
Cache-PW & CrossEncoder & Common & 71/169 & 49.30 [37.88, 61.43] & 0.00 [0.00, 0.00] & +14.58 [10.00, 19.58] \\
Cache-PW & CrossEncoder & World & 358/272 & 44.41 [38.04, 50.99] & 1.47 [0.34, 3.10] & +24.60 [20.48, 29.05] \\
Cache-Sh. & CrossEncoder & Common & 71/169 & 50.70 [38.89, 62.82] & 0.00 [0.00, 0.00] & +15.00 [10.00, 20.00] \\
Cache-Sh. & CrossEncoder & World & 358/272 & 43.30 [37.01, 49.71] & 6.62 [3.64, 10.00] & +21.75 [17.30, 26.35] \\
Cache-PW & Provence & Common & 70/170 & 88.57 [81.82, 95.00] & 4.12 [1.60, 7.27] & +22.92 [16.25, 30.00] \\
Cache-PW & Provence & World & 362/268 & 65.47 [60.16, 70.64] & 22.01 [16.73, 27.59] & +28.25 [23.02, 33.33] \\
Cache-Sh. & Provence & Common & 70/170 & 77.14 [68.06, 86.30] & 7.06 [2.76, 12.57] & +17.50 [10.42, 24.17] \\
Cache-Sh. & Provence & World & 362/268 & 55.25 [49.36, 61.26] & 25.75 [19.62, 32.44] & +20.79 [15.24, 26.19] \\
\bottomrule
\end{tabular*}
\end{table}

\endgroup

\begin{figure}[h]
\centering
\includegraphics[width=\linewidth]{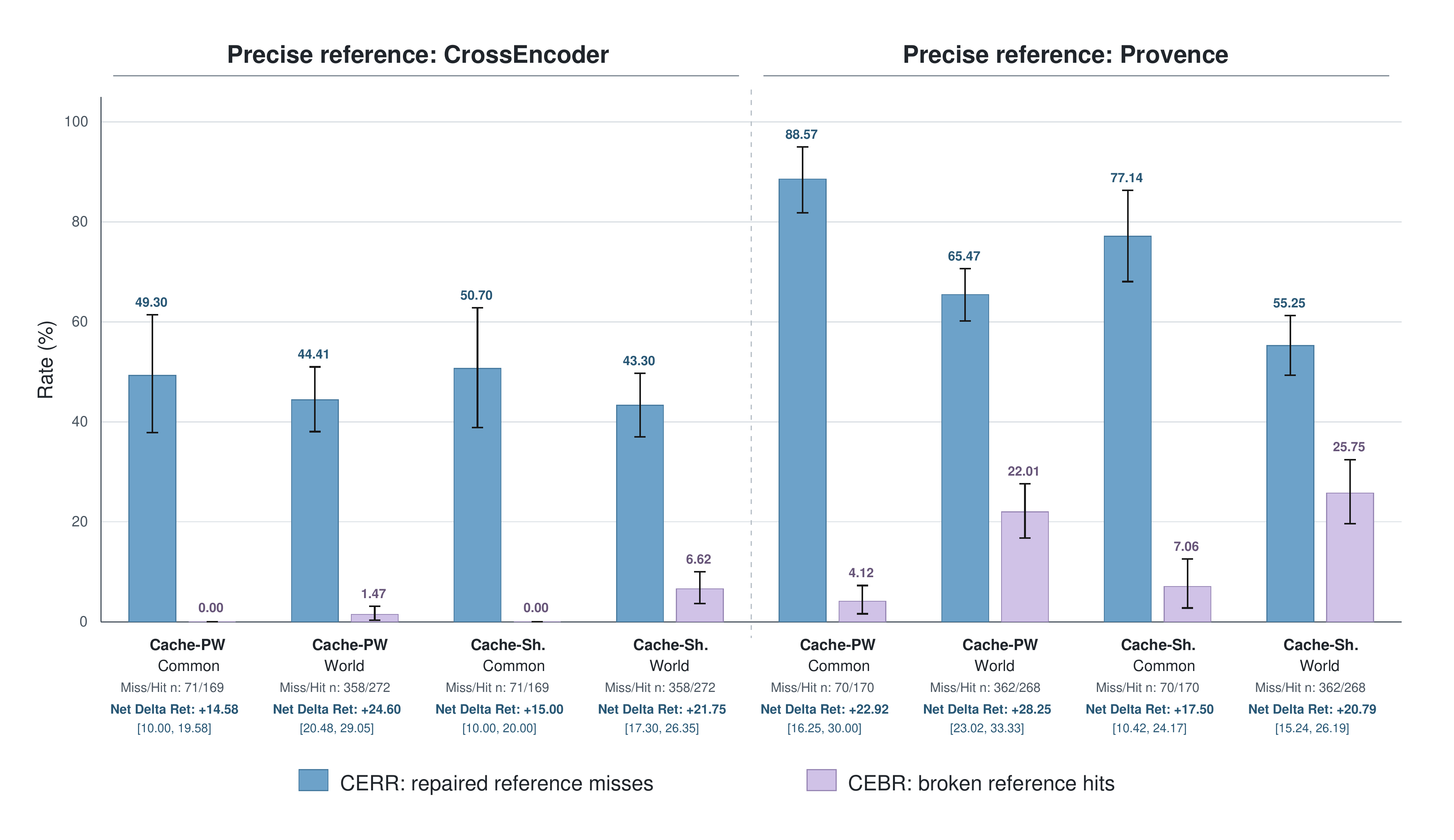}
\caption{NoLiMa evidence-transition rates by knowledge type. Blue bars show CERR, the fraction of precise-only reference misses repaired by the cascade; purple bars show CEBR, the fraction of reference hits broken by the cascade. Error bars denote 95\% haystack--test cluster-bootstrap intervals. Labels below each group report the Miss/Hit denominator and the resulting net retention contrast.}
\label{fig:evidence_transition_rates}
\end{figure}

\begin{table}[H]
\centering
\small
\setlength{\tabcolsep}{7pt}
\caption{Unified NoLiMa component timing in milliseconds. S1 and S2 are mean coarse- and precise-stage times; p50 and p95 cover the complete synchronized online pruning boundary. Cacheable semantic methods store 1,536 bytes per sentence; BM25 has no dense-vector cache.}
\label{tab:cascade_stage_timing}
\begin{tabular}{llcccc}
\toprule
\textbf{Precise} & \textbf{Coarse} & \textbf{S1 mean} & \textbf{S2 mean} & \textbf{p50} & \textbf{p95} \\
\midrule
\multirow{4}{*}{CrossEncoder}
& BM25 & 13.7 & 124.4 & 152.1 & 173.9 \\
& MiniLM & 10.6 & 162.7 & 183.0 & 245.5 \\
& Cache-PW & 16.3 & 119.3 & \textbf{149.6} & \textbf{166.2} \\
& Cache-Sh. & 16.5 & 120.5 & 151.3 & 166.9 \\
\midrule
\multirow{4}{*}{Provence}
& BM25 & 14.7 & 433.6 & 482.0 & 532.4 \\
& MiniLM & 10.9 & 418.6 & 456.5 & 484.8 \\
& Cache-PW & 16.9 & 383.9 & 428.5 & 448.1 \\
& Cache-Sh. & 17.0 & 378.9 & \textbf{421.9} & \textbf{445.3} \\
\bottomrule
\end{tabular}
\end{table}

The 10.6--17.0 ms first-stage means are small relative to the precise stage. The speedup arises from reducing the candidate context processed by the precise reranker, after accounting for online query encoding, cached-feature fusion, selection, and context construction. The 1,536-byte sentence cache equals 1.536 GB per million sentences in decimal units. The reported aggregate latency includes lazy cache construction on first use; warm-cache latency measures reuse of those stored embeddings.

\begin{table}[H]
\centering
\small
\setlength{\tabcolsep}{7pt}
\caption{CrossEncoder coarse-load sensitivity on the same 870 NoLiMa contexts. The final token cap remains 30\%. These measurements come from one matched operating-point sweep; the unified 50\% rerun in Table~\ref{tab:nolima_cascade} supplies the headline timing.}
\label{tab:coarse_load_sensitivity}
\begin{tabular}{lcccc}
\toprule
\textbf{Coarse objective} & \textbf{Admitted} & \textbf{Retention} & \textbf{p50 ms} & \textbf{p95 ms} \\
\midrule
Cache-Pointwise & 50\% & \textbf{0.7253} & \textbf{150.4} & \textbf{161.9} \\
Cache-Pointwise & 70\% & 0.6276 & 191.0 & 205.6 \\
Cache-Shapley & 50\% & \textbf{0.7057} & \textbf{147.1} & \textbf{159.6} \\
Cache-Shapley & 70\% & 0.6333 & 187.6 & 201.7 \\
\bottomrule
\end{tabular}
\end{table}

Sending more candidates to CrossEncoder increases precise-stage mean time from 119.1 to 161.5 ms for Cache-Pointwise and from 116.4 to 158.3 ms for Cache-Shapley. Final evidence retention simultaneously falls by 9.77 and 7.24 points. The precise stage can reorder an enlarged candidate set in a way that removes evidence under the same final token cap. The reported operating point uses the fixed 50\% load; we do not optimize this budget.

Across the two tested loads, admitting fewer candidates to CrossEncoder improves both final evidence retention and online pruning latency.

\subsection{Complete Five-Haystack NoLiMa Results}\label{sec:nolima_books_supp}

\begin{table}[H]
\centering
\scriptsize
\setlength{\tabcolsep}{8pt}
\caption{Complete NoLiMa 32K results by book with Gemma 3 12B. Each entry is answer accuracy / exact evidence retention in percent under the matched 30\% token cap; exact sentence retention is undefined for LLMLingua2's token-compressed text and marked ``--''. Sh., PW, CE, and Prov. abbreviate Shapley, pointwise, CrossEncoder, and Provence. All rows use the same 58 tests, five books, and three insertion depths.}
\label{tab:nolima_books_full_new}
\begin{tabular*}{\linewidth}{lccccccc@{}}
\toprule
\textbf{Method} & \textbf{Book 1} & \textbf{Book 2} & \textbf{Book 3} & \textbf{Book 4} & \textbf{Book 5} & \textbf{All} & \textbf{Tok.} \\
\midrule
\multicolumn{8}{c}{\textit{Baselines and precise rerankers}} \\[2.5pt]
Random & 11.49/24.71 & 13.79/29.89 & 12.64/29.31 & 15.52/28.74 & 14.94/29.89 & 13.68/28.51 & 28.41 \\
BM25 & 12.07/33.33 & 13.79/31.61 & 14.37/24.14 & 16.09/31.03 & 21.26/37.93 & 15.52/31.61 & 28.69 \\
LLMLingua2 & 21.26/-- & 16.67/-- & 13.22/-- & 13.79/-- & 15.52/-- & 16.09/-- & 27.84 \\
CrossEncoder & 21.84/52.87 & 22.41/54.02 & 16.67/44.83 & 26.44/60.92 & 20.11/40.80 & 21.49/50.69 & 27.52 \\
Provence & 22.99/39.08 & 20.11/42.53 & 20.11/43.10 & 26.44/85.63 & 21.84/41.38 & 22.30/50.34 & 28.40 \\
\midrule
\multicolumn{8}{c}{\textit{Cacheable direct rerankers and cascades}} \\[2.5pt]
Frozen MiniLM & 25.29/73.56 & 28.16/70.11 & 27.01/72.41 & 24.71/68.97 & 33.33/75.86 & 27.70/72.18 & 28.50 \\
Cache-Sh. & 31.61/87.36 & 33.33/67.24 & 27.59/69.54 & 29.89/75.86 & 29.31/82.18 & 30.34/76.44 & 28.44 \\
Cache-PW & 33.33/94.83 & 31.61/83.91 & 28.16/82.18 & 28.74/81.03 & 35.63/88.51 & 31.49/86.09 & 28.49 \\
Cache-Sh.$\rightarrow$CE & 29.31/81.61 & 21.84/66.09 & 21.84/63.79 & 26.44/74.14 & 28.16/67.24 & 25.52/70.57 & 28.12 \\
Cache-PW$\rightarrow$CE & 31.61/83.33 & 32.18/75.29 & 24.14/66.67 & 29.31/77.01 & 25.29/60.34 & 28.51/72.53 & 28.14 \\
Cache-Sh.$\rightarrow$Prov. & 33.33/78.74 & 26.44/62.07 & 26.44/67.24 & 25.29/75.29 & 33.33/67.82 & 28.97/70.23 & 28.42 \\
Cache-PW$\rightarrow$Prov. & 34.48/83.91 & 29.89/74.14 & 32.18/71.84 & 28.74/83.33 & 33.33/72.41 & 31.72/77.13 & 28.45 \\
\bottomrule
\end{tabular*}
\end{table}

The book-level results separate isolated outliers from systematic behavior. Direct Provence reaches 85.63\% retention on Book 4 and 39.08--43.10\% on four other books. Frozen MiniLM already provides a strong semantic control at 27.70\% aggregate QA and 72.18\% retention. Cache-Pointwise improves its QA point estimate on every book by 1.15--8.04 points, retains 81.03--94.83\% of exact evidence across the five books, and supplies the highest aggregate direct result. Cache-Shapley has higher direct QA than Cache-Pointwise on Books 2 and 4; Cache-Pointwise leads on Books 1, 3, and 5 and retains more evidence on every book. Its Provence cascade raises aggregate QA to 31.72\% and gains on Books 1 and 3, leaves Book 4 unchanged, and loses on Books 2 and 5 relative to direct Cache-Pointwise. The exact ranking and reader outcomes depend on the haystack and the second-stage reranker. Across these variations, the cacheable coarse stage provides broad evidence coverage.

\begin{table}[H]
\centering
\scriptsize
\setlength{\tabcolsep}{8pt}
\caption{Complete NoLiMa 32K results by book with Llama 3.1 8B as reader. Each entry is answer accuracy / exact evidence retention in percent under the same 30\% token cap; extractive methods reuse the Gemma selections. LLMLingua2 retention is undefined because it compresses within sentences. Tok. is the realized reader-context token ratio.}
\label{tab:nolima_books_llama31_8b}
\begin{tabular*}{\linewidth}{lccccccc@{}}
\toprule
\textbf{Method} & \textbf{Book 1} & \textbf{Book 2} & \textbf{Book 3} & \textbf{Book 4} & \textbf{Book 5} & \textbf{All} & \textbf{Tok.} \\
\midrule
Random & 9.77/24.71 & 12.64/29.89 & 8.62/29.31 & 10.92/28.74 & 14.94/29.89 & 11.38/28.51 & 28.41 \\
BM25 & 4.60/33.33 & 5.75/31.61 & 4.60/24.14 & 6.32/31.03 & 10.92/37.93 & 6.44/31.61 & 28.69 \\
LLMLingua2 & 15.52/-- & 9.77/-- & 17.24/-- & 7.47/-- & 13.79/-- & 12.76/-- & 27.84 \\
CrossEncoder & 21.26/52.87 & 19.54/54.02 & 14.94/44.83 & \textbf{24.71}/60.92 & 17.82/40.80 & 19.66/50.69 & 27.52 \\
Provence & 13.22/39.08 & 16.67/42.53 & 16.09/43.10 & 18.97/85.63 & 17.24/41.38 & 16.44/50.34 & 28.40 \\
Frozen MiniLM & 24.71/73.56 & 22.99/70.11 & 17.24/72.41 & 13.79/68.97 & 28.74/75.86 & 21.49/72.18 & 28.50 \\
Cache-Sh. & 26.44/87.36 & 21.26/67.24 & 13.22/69.54 & 17.24/75.86 & 25.29/82.18 & 20.69/76.44 & 28.44 \\
Cache-PW & \textbf{33.33}/94.83 & \textbf{28.74}/83.91 & 19.54/82.18 & 18.97/81.03 & \textbf{32.18}/88.51 & \textbf{26.55}/86.09 & 28.49 \\
Cache-Sh.$\rightarrow$CE & 24.14/81.61 & 18.97/66.09 & 17.24/63.79 & 19.54/74.14 & 26.44/67.24 & 21.26/70.57 & 28.12 \\
Cache-PW$\rightarrow$CE & 31.03/83.33 & 24.14/75.29 & \textbf{21.84}/66.67 & 20.11/77.01 & 25.29/60.34 & 24.48/72.53 & 28.14 \\
Cache-Sh.$\rightarrow$Prov. & 21.26/78.74 & 16.67/62.07 & 15.52/67.24 & 20.11/75.29 & 28.16/67.82 & 20.34/70.23 & 28.42 \\
Cache-PW$\rightarrow$Prov. & 24.14/83.91 & 20.69/74.14 & 13.79/71.84 & 20.69/83.33 & 29.89/72.41 & 21.84/77.13 & 28.45 \\
\bottomrule
\end{tabular*}
\end{table}

The reader substitution preserves the central coarse-stage result. Direct Cache-Pointwise remains the strongest aggregate method and improves over frozen MiniLM by 5.06 points [2.41, 7.70]. Its CrossEncoder and Provence cascades improve their direct precise stages by 4.83 [1.95, 7.93] and 5.40 points [2.64, 8.16], respectively. Relative to direct Cache-Pointwise, CrossEncoder changes QA by $-2.07$ points [$-4.94$, 0.80], and Provence by $-4.71$ [$-7.47$, $-1.95$]. Book 3 favors the CrossEncoder cascade and Book 4 favors direct CrossEncoder; Cache-Pointwise leads on Books 1, 2, and 5. Cross-reader agreement therefore supports the coarse scorer, and cross-reader and cross-book variation localizes compatibility to the downstream pairing.

\subsection{Complete Question-Type Results}\label{sec:knowledge_slices_supp}

\begin{table}[H]
\centering
\small
\setlength{\tabcolsep}{7pt}
\caption{Cacheable NoLiMa results by question structure with Gemma 3 12B. Each cell is answer accuracy / exact evidence retention in percent. One-hop, two-hop, and two-hop2 contain 480, 300, and 90 outcomes.}
\label{tab:cache_question_types}
\begin{tabular}{lccc}
\toprule
\textbf{Method} & \textbf{One-hop} & \textbf{Two-hop} & \textbf{Two-hop2} \\
\midrule
Cache-Shapley & 38.13 / 83.75 & 22.00 / 79.33 & 16.67 / 27.78 \\
Cache-Pointwise & \textbf{39.17 / 93.33} & \textbf{23.00 / 86.00} & \textbf{18.89 / 47.78} \\
Cache-Sh.$\rightarrow$CE & 33.54 / 80.83 & 15.67 / 63.67 & 15.56 / 38.89 \\
Cache-PW$\rightarrow$CE & \textbf{37.92 / 84.58} & \textbf{16.67} / 61.67 & \textbf{17.78 / 44.44} \\
Cache-Sh.$\rightarrow$Prov. & 37.29 / 78.33 & 19.67 / 68.00 & 15.56 / 34.44 \\
Cache-PW$\rightarrow$Prov. & \textbf{41.46 / 83.96} & \textbf{20.67 / 73.67} & \textbf{16.67 / 52.22} \\
\bottomrule
\end{tabular}
\end{table}

Cache-Pointwise retains more evidence than Cache-Shapley for every question structure. The margin is largest on two-hop2, where direct retention rises from 27.78\% to 47.78\%. After Cache-Pointwise, Provence raises one-hop answer accuracy from 39.17\% to 41.46\%, lowers two-hop accuracy from 23.00\% to 20.67\%, and retains fewer needles than the direct coarse reranker. Question structure therefore exposes the same division observed by knowledge type: the coarse objective controls coverage, and the precise stage reshapes the surviving context for the reader.

\subsection{Additional 2WikiMultiHopQA Paired Statistics}\label{sec:2wiki_paired_supp}

Cache-Pointwise followed by Provence improves F1 by 0.0411 [0.0037, 0.0780] and evidence recall by 0.0771 [0.0512, 0.1025] relative to direct Provence. Cache-Shapley improves evidence recall by 0.0842 [0.0600, 0.1087]. Relative to MiniLM followed by Provence, Cache-Pointwise improves evidence recall by 0.0250 [0.0083, 0.0421], and Cache-Shapley improves it by 0.0321 [0.0138, 0.0500].

For CrossEncoder, MiniLM, Cache-Pointwise, and Cache-Shapley all have F1 intervals that overlap direct CrossEncoder. BM25 reduces F1 by 0.0532 [$-0.0837$, $-0.0234$] and evidence recall by 0.1125 [$-0.1317$, $-0.0933$]. The mean selected-set Jaccard with direct CrossEncoder is 0.871 for MiniLM, 0.855 for Cache-Pointwise, and 0.845 for Cache-Shapley.

Semantic coarse selection therefore improves the Provence evidence set, while it does not significantly change CrossEncoder F1 at the tested budget.

\subsection{Evidence-Fixed Reader Diagnostic}\label{sec:evidence_fixed_supp}

\begin{table}[H]
\centering
\small
\caption{HotpotQA reader diagnostic on 612 held-out examples. Every selector receives all annotated evidence; distractors are selected under a 50\% token cap. Three examples whose evidence alone exceeds the cap retain evidence only. Across selectors, only the selected distractors differ. $\Delta$ is answer accuracy relative to evidence-only context, with a paired 95\% bootstrap interval. QC denotes the query-conditioned SCP models from \citet{chen2026shapleycontextpruningcooperative}.}
\label{tab:evidence_fixed}
\begin{tabular}{lcc}
\toprule
\textbf{Context construction} & \textbf{Accuracy (\%)} & \textbf{$\Delta$ [95\% CI]} \\
\midrule
Evidence only & \textbf{67.65} & 0.00 \\
Random distractors & 62.25 & $-5.39$ [$-8.01$, $-2.78$] \\
BM25 distractors & 62.58 & $-5.07$ [$-7.84$, $-2.29$] \\
MiniLM distractors & 61.76 & $-5.88$ [$-8.99$, $-2.94$] \\
QC-SCP-Shapley distractors & 62.09 & $-5.56$ [$-8.50$, $-2.78$] \\
QC-SCP-Singleton distractors & 63.24 & $-4.41$ [$-7.03$, $-1.80$] \\
QC-Pointwise distractors & 63.40 & $-4.25$ [$-7.03$, $-1.63$] \\
CrossEncoder distractors & 61.76 & $-5.88$ [$-8.82$, $-3.10$] \\
Provence distractors & 62.25 & $-5.39$ [$-8.01$, $-2.78$] \\
\bottomrule
\end{tabular}
\end{table}

The control changes reader input after fixing evidence recall. Every distractor-augmented context lowers answer accuracy relative to evidence alone, establishing a 4.25--5.88 point distractor cost in this protocol. QC-Pointwise and QC-SCP-Singleton differ by 0.16 points [$-1.80$, 2.12], so selector choice has no significant effect in this diagnostic. The experiment establishes the gap between fixed evidence recall and reader accuracy.

\section{Supplementary Experiments}\label{app:supp_experiments}

\subsection{Late-Interaction Endpoint and Three-Stage Cascades}\label{sec:colbert_supp}

\begin{table}[H]
\centering
\small
\setlength{\tabcolsep}{4pt}
\caption{ColBERTv2 endpoint and depth-extension panel on NoLiMa 32K. Speedup divides the direct endpoint's mean latency by each composition's mean latency. Direct ColBERTv2 comes from this panel; the CrossEncoder and Provence references use the unified timing run in Table~\ref{tab:nolima_cascade}. Latency includes cache-miss document-token encoding and online selection. Ret. is exact evidence retention under the common 30\% token cap. PW and Sh. abbreviate Cache-Pointwise and Cache-Shapley.}
\label{tab:colbert_selection_panel}
\begin{tabular}{lcccccc}
\toprule
\textbf{Composition} & \textbf{Fusion} & \textbf{Col. load} & \textbf{$h$ load} & \textbf{p50 ms} & \textbf{Speedup} & \textbf{Ret.} \\
\midrule
\multicolumn{7}{c}{\textit{ColBERTv2 as the final endpoint}} \\ [2pt]
Direct ColBERTv2 & -- & 100\% & -- & 97.7 & 1.00$\times$ & 0.7897 \\
Random$_{50}\rightarrow$ColBERTv2 & Hard & 50\% & -- & 59.9 & 1.71$\times$ & 0.4701 \\
BM25$_{50}\rightarrow$ColBERTv2 & Hard & 50\% & -- & 72.6 & 1.40$\times$ & 0.5805 \\
MiniLM$_{50}\rightarrow$ColBERTv2 & Hard & 50\% & -- & 68.0 & 1.43$\times$ & 0.8149 \\
PW$_{50}\rightarrow$ColBERTv2 & Hard & 50\% & -- & 83.8 & 1.18$\times$ & 0.9149 \\
Sh.$_{50}\rightarrow$ColBERTv2 & Hard & 50\% & -- & 72.6 & 1.35$\times$ & 0.8391 \\
PW$_{70}\rightarrow$ColBERTv2 & Hard & 70\% & -- & 87.6 & 1.13$\times$ & 0.8667 \\
Sh.$_{70}\rightarrow$ColBERTv2 & Hard & 70\% & -- & 87.7 & 1.13$\times$ & 0.8460 \\
PW$_{50}\rightarrow$ColBERTv2 & RRF & 50\% & -- & 72.7 & 1.36$\times$ & \textbf{0.9230} \\
Sh.$_{50}\rightarrow$ColBERTv2 & RRF & 50\% & -- & 73.1 & 1.35$\times$ & 0.8529 \\
\midrule
\multicolumn{7}{c}{\textit{ColBERTv2 before a joint precise reranker}} \\[2pt]
ColBERTv2$_{50}\rightarrow$CrossEncoder & Hard & 100\% & 50\% & 212.7 & 1.12$\times$ & 0.6690 \\
PW$_{70}\rightarrow$ColBERTv2$_{50}\rightarrow$CrossEncoder & Hard & 70\% & 50\% & 203.2 & 1.18$\times$ & 0.6885 \\
Sh.$_{70}\rightarrow$ColBERTv2$_{50}\rightarrow$CrossEncoder & Hard & 70\% & 50\% & 203.1 & 1.17$\times$ & 0.6977 \\
ColBERTv2$_{50}\rightarrow$Provence & Hard & 100\% & 50\% & 505.3 & 1.57$\times$ & 0.7138 \\
PW$_{70}\rightarrow$ColBERTv2$_{50}\rightarrow$Provence & Hard & 70\% & 50\% & 497.4 & 1.59$\times$ & 0.7437 \\
Sh.$_{70}\rightarrow$ColBERTv2$_{50}\rightarrow$Provence & Hard & 70\% & 50\% & 493.6 & 1.62$\times$ & 0.7310 \\
\bottomrule
\end{tabular}
\end{table}

ColBERTv2 supplies a multi-vector semantic interface between reusable single-vector scoring and joint query--candidate encoding. Table~\ref{tab:colbert_selection_panel} evaluates it first as the final reranking endpoint and then as a middle stage. Every row uses the same 870 NoLiMa contexts and 30\% final token cap. Hard cascades retain the final-stage ordering within the admitted set; RRF combines the upstream and ColBERTv2 ranks with $k=60$. The percentages under ``Col. load'' and ``$h$ load'' are fractions of the original sentence pool scored by ColBERTv2 and the final CrossEncoder or Provence stage.

\begin{table}[H]
\centering
\small
\setlength{\tabcolsep}{7pt}
\caption{Paired ColBERTv2 endpoint contrasts on NoLiMa 32K. Entries are percentage-point differences [95\% haystack--test cluster-bootstrap interval] over the same 870 outcomes.}
\label{tab:colbert_endpoint_ci}
\begin{tabular}{lcc}
\toprule
\textbf{Comparison} & $\boldsymbol{\Delta}$\textbf{Retention [95\% CI]} & $\boldsymbol{\Delta}$\textbf{QA [95\% CI]} \\
\midrule
PW Hard $-$ Direct ColBERTv2 & +12.53 [9.31, 15.75] & +4.60 [2.07, 7.24] \\
PW RRF $-$ Direct ColBERTv2 & +13.33 [9.77, 16.90] & +4.60 [1.95, 7.24] \\
Sh. RRF $-$ Direct ColBERTv2 & +6.32 [2.30, 10.34] & +4.14 [1.26, 7.01] \\
Sh. RRF $-$ Sh. Hard & +1.38 [$-0.34$, 3.22] & +1.38 [$-0.34$, 3.10] \\
PW RRF $-$ PW Hard & +0.80 [$-1.15$, 2.76] & 0.00 [$-1.95$, 2.07] \\
Sh. RRF $-$ PW RRF & $-7.01$ [$-9.31$, $-4.83$] & $-0.46$ [$-2.76$, 1.72] \\
\bottomrule
\end{tabular}
\end{table}

Table~\ref{tab:colbert_stage_removal} organizes each row by the removed stage and exposes every directional contrast. Each three-stage Cache-Pointwise chain has three matched references: ColBERTv2$_{50}\rightarrow h$ removes Cache-Pointwise; Cache-Pointwise$_{50}\rightarrow h$ removes ColBERTv2 while preserving the 50\% endpoint load; and Cache-Pointwise$_{70}\rightarrow$ColBERTv2 removes the endpoint while preserving the 70\% ColBERTv2 load. Repeating each chain outcome across these rows makes every directional contrast explicit.

\begin{table}[H]
\centering
\scriptsize
\setlength{\tabcolsep}{3.2pt}
\caption{Key Gemma outcomes and matched stage-removal contrasts for the ColBERTv2 extension. Ret./QA are the chain's absolute exact retention and answer accuracy. Each $\Delta$ subtracts the named removal reference from the chain. All systems use the same 870 examples and 30\% final token cap.}
\label{tab:colbert_stage_removal}
\begin{tabular*}{\textwidth}{@{\extracolsep{\fill}}lllcccc@{}}
\toprule
\textbf{Evaluated chain} & \textbf{Removed} & \textbf{Reference chain} & \textbf{Ret.} & \textbf{QA} & $\boldsymbol{\Delta}$\textbf{Ret.} & $\boldsymbol{\Delta}$\textbf{QA} \\
\midrule
Direct ColBERTv2 & -- & -- & 0.7897 & 0.2954 & -- & -- \\
PW$_{50}\rightarrow$ColBERTv2 & PW & Direct ColBERTv2 & 0.9149 & 0.3414 & +0.1253 & +0.0460 \\
PW$_{70}\rightarrow$ColBERTv2 & -- & -- & 0.8667 & 0.3310 & -- & -- \\
ColBERTv2$_{50}\rightarrow$CrossEncoder & ColBERTv2 & Direct CrossEncoder & 0.6690 & 0.2724 & +0.1621 & +0.0575 \\
ColBERTv2$_{50}\rightarrow$CrossEncoder & CrossEncoder & Direct ColBERTv2 & 0.6690 & 0.2724 & $-0.1207$ & $-0.0230$ \\
PW$_{70}\rightarrow$ColBERTv2$_{50}\rightarrow$CrossEncoder & PW & ColBERTv2$_{50}\rightarrow$CrossEncoder & 0.6885 & 0.2678 & +0.0195 & $-0.0046$ \\
PW$_{70}\rightarrow$ColBERTv2$_{50}\rightarrow$CrossEncoder & ColBERTv2 & PW$_{50}\rightarrow$CrossEncoder & 0.6885 & 0.2678 & $-0.0368$ & $-0.0172$ \\
PW$_{70}\rightarrow$ColBERTv2$_{50}\rightarrow$CrossEncoder & CrossEncoder & PW$_{70}\rightarrow$ColBERTv2 & 0.6885 & 0.2678 & $-0.1782$ & $-0.0632$ \\
\midrule
ColBERTv2$_{50}\rightarrow$Provence & ColBERTv2 & Direct Provence & 0.7138 & 0.2747 & +0.2103 & +0.0517 \\
ColBERTv2$_{50}\rightarrow$Provence & Provence & Direct ColBERTv2 & 0.7138 & 0.2747 & $-0.0759$ & $-0.0207$ \\
PW$_{70}\rightarrow$ColBERTv2$_{50}\rightarrow$Provence & PW & ColBERTv2$_{50}\rightarrow$Provence & 0.7437 & 0.3218 & +0.0299 & +0.0471 \\
PW$_{70}\rightarrow$ColBERTv2$_{50}\rightarrow$Provence & ColBERTv2 & PW$_{50}\rightarrow$Provence & 0.7437 & 0.3218 & $-0.0276$ & +0.0046 \\
PW$_{70}\rightarrow$ColBERTv2$_{50}\rightarrow$Provence & Provence & PW$_{70}\rightarrow$ColBERTv2 & 0.7437 & 0.3218 & $-0.1230$ & $-0.0092$ \\
\bottomrule
\end{tabular*}
\end{table}

\begin{table}[H]
\centering
\small
\setlength{\tabcolsep}{3.5pt}
\caption{Key paired depth-extension contrasts on NoLiMa 32K. Entries are percentage-point differences [95\% haystack--test cluster-bootstrap interval] over the same 870 outcomes.}
\label{tab:colbert_depth_ci}
\begin{tabular*}{\textwidth}{@{\extracolsep{\fill}}lcc@{}}
\toprule
\textbf{Comparison} & $\boldsymbol{\Delta}$\textbf{Retention [95\% CI]} & $\boldsymbol{\Delta}$\textbf{QA [95\% CI]} \\
\midrule
ColBERTv2$_{50}\rightarrow$CE $-$ Direct CE & +16.21 [13.10, 19.43] & +5.75 [3.33, 8.16] \\
PW$_{70}\rightarrow$ColBERTv2$_{50}\rightarrow$CE $-$ ColBERTv2$_{50}\rightarrow$CE & +1.95 [0.92, 3.10] & $-0.46$ [$-2.41$, 1.49] \\
ColBERTv2$_{50}\rightarrow$Prov. $-$ Direct Prov. & +21.03 [16.78, 25.29] & +5.17 [2.30, 8.05] \\
PW$_{70}\rightarrow$ColBERTv2$_{50}\rightarrow$Prov. $-$ ColBERTv2$_{50}\rightarrow$Prov. & +2.99 [$-0.46$, 6.67] & +4.71 [1.95, 7.47] \\
PW$_{70}\rightarrow$ColBERTv2$_{50}\rightarrow$Prov. $-$ PW$_{50}\rightarrow$Prov. & $-2.76$ [$-6.78$, 1.26] & +0.46 [$-2.30$, 3.22] \\
\bottomrule
\end{tabular*}
\end{table}

Learned coarse selection improves direct ColBERTv2, whereas the benefit of an additional stage depends on the downstream reranker and the chain used as reference.

\subsection{Objective, Readout, and Multi-Seed Diagnostics}\label{sec:mc_noise_supp}\label{sec:mc_robustness}

\begingroup
\setlength{\intextsep}{7pt}

Cache-Pointwise and Cache-Shapley share representations and singleton deployment and differ in independent versus sampled-coalition training credit. Table~\ref{tab:mc_noise} measures the Monte-Carlo credit noise of the Shapley-trained scorer on 256 validation contexts: credit standard deviation scales close to $1/\sqrt{M}$, the sign-disagreement rate of pairwise margins falls from 2.43\% to 0.46\%, and the rank correlation to the high-budget reference credit exceeds 0.97 already at $M{=}1$. The estimator supplies effective gradients at the tested training budgets (Table~\ref{tab:mtrain_sweep}). Deployment averaging over more permutations dilutes exact-needle ranking (Table~\ref{tab:readout_paired}), making linear-time singleton scoring the strongest tested readout. Across three seeds, Pointwise achieves the higher mean and peak retention; Shapley has the higher minimum and smaller range (Table~\ref{tab:multiseed_retention}). These diagnostics separate the effects of training credit from the cost and quality of deployment readout.

\begin{table}[H]
\centering
\small
\setlength{\tabcolsep}{7pt}
\caption{Monte-Carlo credit-noise diagnostics for the Shapley-trained scorer (256 validation contexts). ``Sign disagreement'' is the fraction of positive--negative margin flips relative to the high-budget reference; the last column is the Spearman correlation to that reference credit.}
\label{tab:mc_noise}
\begin{tabular}{ccccc}
\toprule
\textbf{$M$} & \textbf{Credit SD} & \textbf{Margin SD} & \textbf{Sign disagreement (\%)} & \textbf{Spearman to ref.} \\
\midrule
1 & 0.0181 & 0.0265 & 2.43 & 0.9752 \\
5 & 0.0081 & 0.0118 & 1.16 & 0.9899 \\
10 & 0.0057 & 0.0083 & 0.88 & 0.9935 \\
25 & 0.0036 & 0.0052 & 0.59 & 0.9958 \\
50 & 0.0025 & 0.0037 & 0.46 & 0.9971 \\
\bottomrule
\end{tabular}
\end{table}

\begin{table}[H]
\centering
\small
\setlength{\tabcolsep}{7pt}
\caption{Training-time Monte-Carlo budget sweep (NPRS). All checkpoints share the encoder, data, and optimization budget and are evaluated on the same frozen embeddings. The budget curve is non-monotonic; MAP shows the same pattern.}
\label{tab:mtrain_sweep}
\begin{tabular}{lcccc}
\toprule
\textbf{Method} & \textbf{MS MARCO} & \textbf{HotpotQA} & \textbf{MuSiQue} & \textbf{2WikiMH} \\
\midrule
Frozen cosine & 0.8352 & 0.8109 & \textbf{0.7478} & 0.8253 \\
Pointwise & 0.8479 & 0.8321 & 0.7305 & 0.8388 \\
Shapley $M{=}1$ & 0.8476 & \textbf{0.8338} & 0.7290 & \textbf{0.8423} \\
Shapley $M{=}5$ & \textbf{0.8481} & 0.8332 & 0.7296 & 0.8422 \\
Shapley $M{=}10$ & 0.8477 & 0.8327 & 0.7285 & 0.8389 \\
Shapley $M{=}25$ & 0.8480 & 0.8323 & 0.7258 & 0.8417 \\
\bottomrule
\end{tabular}
\end{table}

\begin{table}[H]
\centering
\small
\setlength{\tabcolsep}{7pt}
\caption{NoLiMa 32K exact retention of the training-budget checkpoints (30\% token cap, 870 outcomes). Retention is non-monotonic across $M_{\mathrm{train}}$; $M_{\mathrm{train}}{=}25$ is the best tested seed-42 checkpoint, and standard ranking remains nearly flat (Table~\ref{tab:mtrain_sweep}).}
\label{tab:mtrain_retention}
\begin{tabular}{lc}
\toprule
\textbf{Checkpoint} & \textbf{Retention (\%)} \\
\midrule
Shapley $M{=}1$ & 74.94 \\
Shapley $M{=}5$ & 71.61 \\
Shapley $M{=}10$ & 76.44 \\
Shapley $M{=}25$ & \textbf{79.08} \\
\bottomrule
\end{tabular}
\end{table}

\begin{table}[H]
\centering
\small
\setlength{\tabcolsep}{7pt}
\caption{NoLiMa 32K exact retention under one Shapley checkpoint with varying deployment-time readout (30\% token cap, 870 paired outcomes). Singleton is both the strongest and the cheapest readout; larger Monte-Carlo budgets dilute exact-needle retention.}
\label{tab:readout_sweep}
\begin{tabular}{lccc}
\toprule
\textbf{Readout} & \textbf{Retention (\%)} & \textbf{95\% CI} & \textbf{Scoring (ms)} \\
\midrule
Singleton & \textbf{76.44} & [72.41, 80.23] & \textbf{16.87} \\
$M{=}5$ & 70.23 & [66.55, 73.79] & 20.21 \\
$M{=}10$ & 68.51 & [64.94, 71.84] & 21.51 \\
$M{=}20$ & 62.18 & [58.74, 65.63] & 23.57 \\
$M{=}50$ & 49.77 & [46.32, 53.33] & 30.15 \\
\bottomrule
\end{tabular}
\end{table}

\endgroup

\begin{table}[H]
\centering
\small
\setlength{\tabcolsep}{7pt}
\caption{Paired retention comparisons for the deployment-time readout sweep of Table~\ref{tab:readout_sweep} (870 paired outcomes, cluster bootstrap over haystack--test clusters).}
\label{tab:readout_paired}
\begin{tabular}{lcc}
\toprule
\textbf{Comparison} & \textbf{Retention diff.\ (pt) [95\% CI]} & \textbf{Top-$K$ Jaccard} \\
\midrule
Singleton $-$ $M{=}5$ & $+6.21$ [$+2.87$, $+9.54$] & 0.4905 \\
Singleton $-$ $M{=}10$ & $+7.93$ [$+4.25$, $+11.61$] & 0.4307 \\
Singleton $-$ $M{=}20$ & $+14.25$ [$+10.23$, $+18.16$] & 0.3639 \\
Singleton $-$ $M{=}50$ & $+26.67$ [$+22.18$, $+30.92$] & 0.2793 \\
$M{=}5$ $-$ $M{=}10$ & $+1.72$ [$-0.92$, $+4.37$] & 0.6351 \\
$M{=}10$ $-$ $M{=}20$ & $+6.32$ [$+3.68$, $+8.97$] & 0.5480 \\
$M{=}20$ $-$ $M{=}50$ & $+12.41$ [$+8.97$, $+15.86$] & 0.4053 \\
$M{=}5$ $-$ $M{=}50$ & $+20.46$ [$+16.55$, $+24.25$] & 0.3038 \\
\bottomrule
\end{tabular}
\end{table}

\begin{table}[H]
\centering
\small
\setlength{\tabcolsep}{7pt}
\caption{Per-seed NPRS of the multi-seed cacheable training. Every cell uses the same frozen embeddings and evaluation split.}
\label{tab:multiseed_detail}
\begin{tabular}{llcccc}
\toprule
\textbf{Objective} & \textbf{Seed} & \textbf{MS MARCO} & \textbf{HotpotQA} & \textbf{MuSiQue} & \textbf{2WikiMH} \\
\midrule
Pointwise & 42 & 0.8479 & 0.8321 & 0.7305 & 0.8388 \\
Pointwise & 123 & 0.8476 & 0.8330 & 0.7312 & 0.8411 \\
Pointwise & 2026 & 0.8473 & 0.8281 & 0.7301 & 0.8438 \\
Shapley & 42 & 0.8477 & 0.8327 & 0.7285 & 0.8389 \\
Shapley & 123 & 0.8480 & 0.8313 & 0.7266 & 0.8405 \\
Shapley & 2026 & 0.8475 & 0.8311 & 0.7220 & 0.8404 \\
\bottomrule
\end{tabular}
\end{table}

\begin{table}[H]
\centering
\small
\setlength{\tabcolsep}{7pt}
\caption{NoLiMa 32K exact retention (\%) across three training seeds (30\% token cap, 870 outcomes per seed). Pointwise attains the higher mean and maximum; Shapley has the smaller observed range and higher minimum.}
\label{tab:multiseed_retention}
\begin{tabular}{lcccc}
\toprule
\textbf{Objective} & \textbf{seed 42} & \textbf{seed 123} & \textbf{seed 2026} & \textbf{Range} \\
\midrule
Pointwise & 86.09 & 78.74 & 68.51 & 17.59 \\
Shapley & 76.44 & 72.41 & 73.45 & \textbf{4.02} \\
\bottomrule
\end{tabular}
\end{table}

Singleton remains the strongest tested Shapley readout; Pointwise attains higher mean NoLiMa retention, while Shapley shows less variation across the three training seeds.

\section{Design Justification and Research Outlook}\label{app:theory_future}

This appendix states the computational distinction behind cacheable coarse ranking and records the system and objective questions opened by the experiments.

\subsection{Why Cacheable Coarse Ranking Reduces Precise-Stage Cost}

An independently encoded corpus vector is computed once and reused across queries. Online query encoding and feature fusion scale linearly with the number of candidates. CrossEncoder and Provence process query--candidate token interactions for every candidate they receive. A cacheable coarse stage therefore reduces the number of expensive joint evaluations. Table~\ref{tab:nolima_cascade} measures the complete online pruning boundary and includes the additional first-stage work. A complete computational account tracks indexing cost alongside the measured query encoding, feature scoring, precise-stage candidates, synchronized latency, memory, and reader input.

In the measured cascades, the reduction in precise-stage work exceeds the added coarse-stage cost, producing a net online pruning speedup.

\subsection{Implications for Multistage Pruning}\label{sec:multistage_implications}

Stage-removal comparisons support insertion and removal under an application's quality and cost constraints. Declare $(h,\alpha,B,\mathcal R)$ and compare $h$, $g$, and $g{\rightarrow}h$ on matched examples. For Cache-PW$\rightarrow$CrossEncoder (Table~\ref{tab:stage_signature_main}), the contrast with direct CrossEncoder gives +21.84 retention points, +7.01 QA points, and 1.58$\times$ speed at 0.56~GiB additional memory, favoring the cascade when that memory is feasible. The reverse contrast shows that CrossEncoder reduces direct Cache-PW retention by 13.56 and QA by 2.99 points, favoring coarse-only deployment when the endpoint is optional.

The three-stage panel in Appendix~\ref{sec:colbert_supp} extends this choice to pipeline depth: adding coarse selection has different reader effects before ColBERTv2--CrossEncoder and ColBERTv2--Provence. Stage type, order, candidate load, and rank fusion therefore provide concrete variables for choosing which stages to combine and how much work to assign to each.

\subsection{Research Outlook}\label{sec:research_outlook}

\paragraph{Training credit and deployment readout.} The matched objectives show distinct peak and stability behavior, with singleton deployment as the strongest tested readout. Controlled evidence relations, multi-seed QA, and matched stronger encoders can connect pointwise or coalition-derived training credit to reader outcomes. Sweeps over $M_{\mathrm{train}}$, $M_{\mathrm{readout}}$, evidence density, and context length can identify when a learned set score transfers to singleton ranking.

\paragraph{Objective--representation interaction.} The BGE pointwise control confirms that representation choice changes transfer and the relation between retention and QA. A matched BGE Shapley model and multi-seed cascade evaluation can separate the contribution of the frozen representation from the training-credit objective.

\paragraph{Relation-specific interaction analysis.} Controlled examples with duplicated, complementary, conflicting, and individually insufficient evidence can connect conditional marginal patterns to reader behavior. Interaction indices can then analyze a specified learned game with relation labels and paired reader outcomes.

\paragraph{Multistage pruning and interactions among rerankers.} The quality and speed gains from separately trained stages motivate a broader study of heterogeneous pipelines. Varying stage order, candidate loads, and rank fusion can reveal which scoring mechanisms complement one another and when further pruning removes useful evidence. Our two- and three-stage panels provide starting points for adaptive routing, early exit, and joint quality--latency optimization. Joint training can then be compared with independently trained compositions under the same stage-removal references. Large-corpus experiments should report indexing, cache memory, throughput, synchronized latency, and answer quality together.

\paragraph{Broader long-context evaluation.} FictionLiveBench and other non-literal narrative benchmarks can test transfer beyond NoLiMa. A complete evaluation should cover multiple lengths, token budgets, stories, readers, and insertion positions with cluster-level uncertainty.

These directions aim to identify which pruning stages preserve or improve reader quality at an acceptable computational cost across datasets and serving conditions.

\end{document}